\documentclass[lettersize,journal]{IEEEtran}
\usepackage{amsmath,amsfonts}
\usepackage{algorithmic}
\usepackage{algorithm}
\usepackage{array}
\usepackage{textcomp}
\usepackage{stfloats}
\usepackage{url}
\usepackage{verbatim}
\usepackage{graphicx}
\usepackage{cite}
\usepackage{graphicx}
\usepackage{caption}
\usepackage{subcaption}
\usepackage{booktabs}
\usepackage{tabularx}
\usepackage{diagbox}
\usepackage{multirow}
\usepackage[notransparent]{svg}
\usepackage[colorlinks=true, linkcolor=blue, citecolor=blue, urlcolor=blue]{hyperref}

\usepackage{xcolor}
\usepackage{enumitem}
\usepackage[figuresright]{rotating}
\usepackage{multirow}
\usepackage{soul, color, xcolor}
\usepackage{tabularx} 
\usepackage{booktabs} 
\usepackage{graphicx}
\usepackage{subcaption}
\usepackage{colortbl,xcolor}
\usepackage{wrapfig}
\usepackage{amsmath}
\usepackage{ulem}
\usepackage{amssymb}

\usepackage{bbm}

\begin{document}


\title{Language Models as Messengers: Enhancing Message Passing in Heterophilic Graph Learning}

\author{ Dawei Cheng, 
    Wenjun Wang,
    Mingjian Guang

\thanks{The work is supported by the National Science Foundation of China (62522213).}
\thanks{Dawei Cheng, Wenjun Wang, are with the School of Computer Science and Technology, Tongji University, Shanghai, China.  (e-mail: \{dcheng, 2251275\}@tongji.edu.cn)}
\thanks{Mingjian Guang is with the School of Information and Intelligent Science, Donghua University, Donghua University, Shanghai 201620, China (e-mail: guangmingjian@dhu.edu.cn)}


}



\maketitle

\begin{abstract}
Graph neural networks (GNNs) have become a standard paradigm for graph representation learning, yet their message passing mechanism implicitly assumes that messages can be represented by source node embeddings, an assumption that fails in heterophilic graphs.
While existing methods attempt to address heterophily through graph structure refinement or adaptation of neighbor aggregation, they often overlook the semantic potential of node text, relying on suboptimal message representation for propagation and compromise performance on homophilic graphs.
To address these limitations, we propose LEMP4HG, a novel language model (LM)-enhanced message passing approach for heterophilic graph learning. 
Specifically, for text-attributed graphs (TAG), we leverage a LM to explicitly model inter-node semantic relationships from paired node texts, synthesizing semantically informed messages for propagation. 
To ensure practical efficiency, we further introduce an active learning–inspired strategy guided by a tailored heuristic, MVRD, which selectively enhances messages for node pairs most affected by message passing. 
Extensive experiments demonstrate that LEMP4HG consistently outperforms state-of-the-art methods on heterophilic graphs while maintaining robust performance on homophilic graphs under a practical computational budget.
\end{abstract}


\section{Introduction}
\IEEEPARstart{G}{raph-structured} data, which represent entities and their relationships through nodes and edges, are ubiquitous across diverse real-world domains \cite{frasconi1998general,goller1996learning}. 
To model such data, a wide range of graph neural network (GNN) architectures have been developed, most of which are fundamentally built upon the message passing paradigm, where node representations are updated by aggregating information from neighboring nodes. 
This paradigm implicitly assumes that messages propagated along edges can be directly represented by source node embeddings, i.e., $\textit{message} = \textit{neighbor representation}$. 
While this assumption holds in homophilic graphs \cite{mcpherson2001birds,pei2020geom,hamilton2020graph}, it fails in heterophilic graphs \cite{zhu2020beyond,bo2021beyond,luan2022revisiting,luan2021heterophily}, where neighboring nodes often exhibit dissimilar attributes and labels. 
As a result, naive message propagation introduces semantic conflicts, leading to representation degradation and unstable learning behavior. 
Importantly, this limitation stems from a semantic mismatch in message passing, rather than from graph structure alone.

Existing efforts to address heterophily in GNNs can be broadly categorized into graph structure refinement and GNN architecture adaptation. 
Structure refinement methods aim to adjust the receptive field of each node by incorporating non-local or multi-hop neighbors, or by removing potentially misleading connections, thereby mitigating the adverse effects of heterophilic edges \cite{zou2023se,bi2024make,yang2024graph}. 
In contrast, architecture adaptation methods focus on redesigning message passing and representation update mechanisms to better accommodate heterophilic neighborhoods \cite{song2023ordered,zhang2024unleashing,liu2025integrating}. More recently, several studies have explored incorporating language models into heterophilic graph learning by encoding node texts or guiding edge reweighting \cite{wu-etal-2025-exploring}.

While above approaches alleviate heterophily to some extent, most of them preserve the conventional message passing formulation, leaving the semantic contradiction in heterophilic propagation fundamentally unresolved. 
Moreover, many heterophily-oriented GNNs overlook or underutilize the rich textual attributes widely available in real-world graphs \cite{grave2018learning,wu-etal-2025-exploring}. 
Even when textual information is incorporated, language models are typically used as node-level feature encoders rather than for modeling inter-node semantic relationships, causing messages to remain anchored to source node features and resulting in persistent signal conflicts in heterophilic regions. 
In addition, performance improvements on heterophilic graphs are often achieved at the expense of degraded accuracy on homophilic graphs, limiting the practical robustness of existing methods \cite{zhu2020beyond,chien2020adaptive,lim2021large,he2021bernnet}.

To address the above limitations, we argue that robust learning under heterophily requires rethinking message representation, particularly in text-attributed graphs (TAGs) \cite{fan2024graph,ren2024survey}. 
Rather than treating messages as proxies of source node embeddings, we explicitly model the semantic relationships between paired nodes. 
To this end, we leverage large language models (LMs) to generate connection analyses from paired node texts, capturing relational semantics that cannot be inferred from individual node representations. 
These analyses are encoded and adaptively fused with paired node embeddings via a gating mechanism, yielding discriminative and context-aware messages.
However, applying LM-based message generation to all edges incurs prohibitive $O(E)$ complexity. 
We therefore introduce an active learning–inspired strategy \cite{cai2017active} that selectively enhances messages only for edges where conventional message passing induces significant representation distortion. 
Specifically, we identify such critical edges using a tailored heuristic, termed MVRD, and query LMs exclusively for these informative node pairs. 
This selective enhancement substantially reduces computational overhead while preserving robustness across both heterophilic and homophilic regions.

In summary, our main contributions are as follows: 
\begin{itemize}
    \item We propose LEMP4HG, a novel LM-enhanced message passing approach for heterophilic graph, which encode and fuse LM-generated textual content with paired node texts to obtain enhanced message representations for propagation between connected nodes.
    \item We propose an active learning strategy guided by our heuristic MVRD, selectively enhance paired nodes suffering most from message passing measured by embedding shift, which reduces the cost of textual content generation and side effects on homophilic regions.  
    \item We conduct extensive experiments on 16 real-world datasets, which demonstrate that our LEMP4HG excels on heterophilic graphs and also delivers robust performance on homophilic graphs. 
\end{itemize}

\section{Preliminaries}
\subsection{Problem Definition}
We consider a text-attributed graph $\mathcal{G}^T=(\mathcal{V},\mathcal{E},\mathcal{T})$, where $\mathcal{V}=\{v_1,\dots,v_N\}$ is the node set, $\mathcal{E}$ is the edge set without self-loops, and $\mathcal{T}=\{t_1,\dots,t_N\}$ denotes node-associated texts. The adjacency matrix is $\mathcal{A}\in\mathbb{R}^{N\times N}$, where $A_{ij}=1$ if $(v_i,v_j)\in\mathcal{E}$ and $0$ otherwise. Each node $v_i$ is associated with a textual embedding $\boldsymbol{x}_i$ encoded by an SLM. We focus on semi-supervised transductive node classification on text-attributed graphs.

\subsection{Classic Message Passing Mechanism}
\label{classic MP}
In GNNs, node representations are updated via neighborhood aggregation and state update. At layer $l$, the message passing process for node $v_i$ is defined as:
\begin{equation}
	\boldsymbol{m}_i^l = \sigma\!\left(AGGR(\{\boldsymbol{h}_j^{l-1}\mid v_j\in\mathcal{N}(v_i)\})\right),
\end{equation}
\begin{equation}
	\boldsymbol{h}_i^l = UPDATE(\boldsymbol{h}_i^{l-1}, \boldsymbol{m}_i^l),
\end{equation}
where $\boldsymbol{h}_j^{l-1}$ is the representation of neighbor $v_j$, $\mathcal{N}(\cdot)$ is neighborhood function, $\boldsymbol{m}_i^l$ is aggregated message for $v_i$ and $\boldsymbol{h}_i^l$ is the updated representation of $v_i$ in $l$-th layer. While effective for homophilic graphs, this mechanism may degrade in heterophilic settings where neighboring nodes tend to have dissimilar features or labels.

\subsection{Graph-aware and Graph-agnostic Models}
Neural networks that aggregate neighbors based on graph structure are called graph-aware models, typically paired with a graph-agnostic one. For example, removing the neighboring aggregation from a 2-layer GCN reduces it to a 2-layer MLP.
\begin{equation}
	\sigma\!\left(
	\boldsymbol{\hat{A}}_{sym}
	\sigma(\boldsymbol{\hat{A}}_{sym} \boldsymbol{X} \boldsymbol{W_0})
	\boldsymbol{W_1}
	\right)\\
	\longrightarrow
	\sigma(\sigma(\boldsymbol{X} \boldsymbol{W_0})\boldsymbol{W_1})
\end{equation}
where $\boldsymbol{\hat{A}}_{sym}=\boldsymbol{\tilde{D}}^{-1/2}\boldsymbol{\tilde{A}}\boldsymbol{\tilde{D}}^{-1/2}$, $\boldsymbol{\tilde{A}}\equiv \boldsymbol{A}+\boldsymbol{I}$ and $\boldsymbol{\tilde{D}}\equiv \boldsymbol{D}+\boldsymbol{I}$. In graph $G$, $\boldsymbol{A}$ is adjacency matrix, $\boldsymbol{D}$ is diagonal degree matrix and $\boldsymbol{I}$ is identity matrix, $\sigma$ is activation function.

\subsection{Homophily Metrics}
Homophily characterizes label consistency in graph structures. Two commonly used metrics are edge homophily and node homophily:
\begin{equation}
	\mathcal{H}_{edge}=\frac{|\{e_{ij}\in\mathcal{E}\mid y_i=y_j\}|}{|\mathcal{E}|},
\end{equation}
\begin{equation}
	\mathcal{H}_{node}=\frac{1}{|\mathcal{V}|}\sum_{v_i\in\mathcal{V}}\frac{|\{v_j\in\mathcal{N}(v_i)\mid y_j=y_i\}|}{d_i},
\end{equation}
where $d_i$ is the degree of $v_i$. These metrics quantify global and local label consistency, respectively.

\section{The Proposed Method}
\begin{figure*}
    \centering
    \includegraphics[width=0.98\textwidth]{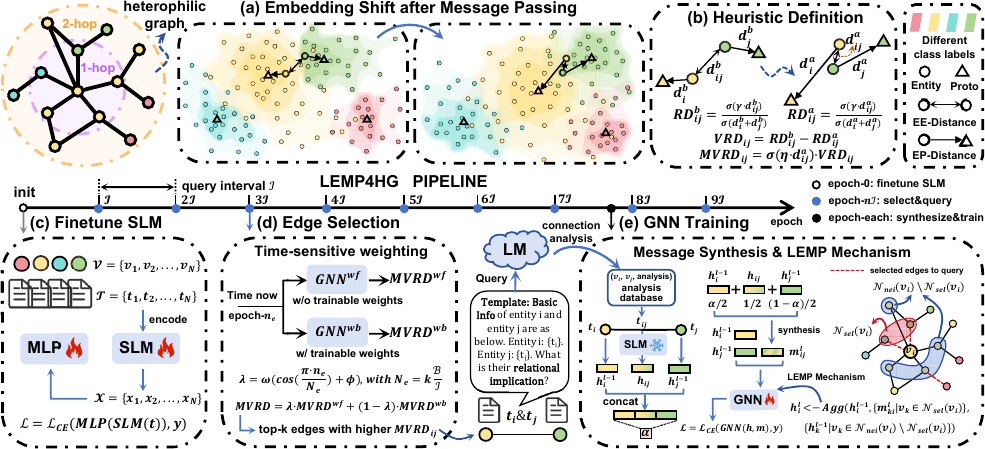}
    \caption{Overview of our LEMP4HG. (a) Illustration of embedding shift after message passing; (b) Heuristic definition to measure how much node pair suffer from message passing. Our pipeline includes three parts. (c) Initially, we finetune SLM for textual encoding with MLP as classifier; (d) Every $\mathcal{I}$ epochs, we select edges by MVRD to query LM for connection analysis; (e) Each epoch, we synthesize all encoded analysis and paired node texts to form enhanced messages for GNN training.}
    \label{fig:framework}
\end{figure*}

Fig. \ref{fig:framework} provides the overview of LEMP4HG. The core idea is to mitigate semantic distortion caused by embedding shift during message passing through LM-enhanced message synthesis. We employ a heuristic-guided edge selection strategy to identify distortion-prone edges for periodic querying, after which the synthesized messages are incorporated into a lightweight GNN for representation learning.
\subsection{LM-Enhanced Message Passing Mechanism}
In heterophilic regions, traditional message passing inevitably fuse contradictory signals between connected nodes, leading to suboptimal patterns learned by model. To address this, we propose a LM-Enhanced Message Passing (LEMP) mechanism, which can be summarized into the following three stages: LM Message Generation, Discriminative Message Synthesis, and Message Passing.

\subsubsection{LLM Message Generation}
We design prompts ${\pi}$ (detailed in Appendix \ref{prompt}) to query LM ${\Psi}_{LM}$ for the connection analysis of node pair $(v_i,v_j)$ with their associated texts $t_i$ and $t_j$. The response $t_{ij}$ is then encoded by finetuned SLM ${\Phi}_{SLM}$ as the preliminary message $\boldsymbol{h}_{ij}$ for the subsequent process.
\begin{equation}
    \boldsymbol{h}_{ij}={\Phi}_{SLM}\circ{\Psi}_{LM}(t_i, t_j;{\pi}),\quad \forall\ e_{ij}\in\mathcal{E}\ \ \text{and}\ \ i\neq j
\end{equation}


\subsubsection{Discriminative Message Synthesis}
To lower the cost of analysis generation by LM, node pair $(v_i,v_j)$ and $(v_j,v_i)$ share the same preliminary message, i.e. $\boldsymbol{h}_{ij}\equiv \boldsymbol{h}_{ji}$. However, the static nature of preliminary messages may hinder long-range neighbor aggregation, which relies on iteratively updated node representations in traditional message passing. Moreover, connection analysis may differ in semantic form from node texts, leading to misalignment in their encoded embeddings and noise introduction. To address these, we introduce a discriminative gating mechanism to fuse preliminary messages with source and target node embeddings, yielding final LM-enhanced messages.
\begin{equation}
\label{concat}
\boldsymbol{\alpha}_{ij}^l = \sigma\left( \left[ \boldsymbol{h}_i^{l-1} \,\|\, \boldsymbol{h}_{ij} \,\|\, \boldsymbol{h}_j^{l-1} \right] \boldsymbol{W}_{gate} \right)
\end{equation}
\begin{equation}\label{gate}
    \boldsymbol{m}_{ij}^{l} = \beta\boldsymbol{h}_{ij} + (1-\beta)\left[\boldsymbol{\alpha}_{ij}^l\odot \boldsymbol{h}_i^{l-1} + (\boldsymbol{1} - \boldsymbol{\alpha}_{ij}^l)\odot \boldsymbol{h}_j^{l-1}\right]
\end{equation}
where $\sigma$ is an activation function (e.g. Sigmoid), and $\boldsymbol{W}_{gate}$ is a trainable weight matrix. For node pair $(v_i,v_j)$, we concatenate $\boldsymbol{h}_i^{l-1}$ and $\boldsymbol{h}_j^{l-1}$ with $\boldsymbol{h}_{ij}$ to compute the gate weight $\boldsymbol{\alpha}_{ij}^l$, which together with a hyperparameter $\beta$ controls their contributions to the fused message $\boldsymbol{m}_{ij}^l$.

\subsubsection{Message Passing}
Unlike the classic message passing mechanism as shown in Preliminary \ref{classic MP}, we employ LM-enhanced message $\boldsymbol{m}_{ij}^{l}$ to substitute the neighbor representation $\boldsymbol{h}_j^{l-1}$ for propagation.
\begin{align}
\boldsymbol{h}_j^{l}
&= UPDATE\!\left(
\boldsymbol{h}_j^{l-1},
\{\boldsymbol{m}_{ij} \mid v_i \in \mathcal{N}(v_j)\}
\right) \notag \\
&= \sigma\!\left(
\boldsymbol{\hat{A}}_{\mathrm{sym}}^{jj}\boldsymbol{h}_j^{l-1}
+ \sum_{v_i \in \mathcal{N}(v_j)}
\boldsymbol{\hat{A}}_{\mathrm{sym}}^{ij}\boldsymbol{m}_{ij}^l
\right)
\end{align}
where $\sigma$ includes batch normalization, activation functions (e.g. ReLU), and dropout.

\subsection{Heuristic for Evaluating Message Passing}
\label{heuristic}
\textbf{Assumptions.} \textit{(1) Nodes with similar features are more likely to share the same category labels; (2) A node’s representation and its classification confidence tend to be more reliable when it lies nearer to its embedding cluster center; (3) GNNs favor mild smoothing, while excessive contraction of representations between heterophilic node pairs usually indicates representation distortion.}


Building on these assumptions, we propose \textbf{MVRD} (\textbf{M}odulated \textbf{V}ariation of \textbf{R}eliable \textbf{D}ifference) as a heuristic to evaluate the effect of message passing from the perspective of paired node embedding contraction, capturing representation distortion commonly arise in heterophilic regions and suppress benign convergence typically in homophilic regions. The specific calculation steps are as below.


\subsubsection{Reliable Difference}
To evaluate the difference between the connected nodes in the embedding space reliably, we firstly cluster the node representations in the embedding space in a semi-supervised way. Specifically, we utilize the labeled $h^l$ to calculate the cluster centers $c_k$ for each class $k\in\{0,1,...,K\}$ and obtain the pseudo labels $\hat{y}_i$ for all the nodes by first estimating the cluster centers as $c_k = \frac{1}{|\{ i \mid \hat{y_i} = k \}|} \sum_{i:\hat{y_i} = k} h_i$ for all $k \in \{0,\ldots,K\}$, and then assigning each node to the nearest center via $\hat{y}_i = \arg\min_k || h_i - c_k ||^2$. We then re-calculate the cluster centers using the obtained pseudo labels $\hat{y}_i$ and all node representations $h_i$ in the same manner. 

For each node pair $(v_i,v_j)$ with $e_{ij}\in\mathcal{E}$, we compute their euclidean distance $d_{ij}$ and their respective distances to cluster centers, $d_i$ and $d_j$. Then, the reliable difference $RD_{ij}$ between node $v_i$ and $v_j$ in the embedding space can be measured as below.
\begin{equation}\label{rd}
    RD_{ij}=\frac{\sigma(\gamma\cdot d_{ij})}{\sigma(d_i+d_j)}=\frac{\sigma(\gamma\cdot \|\boldsymbol{h}_i-\boldsymbol{h}_j\|)}{\sigma(\|\boldsymbol{h}_i-\boldsymbol{c}_{\hat{y}_i}\|+\|\boldsymbol{h}_j-\boldsymbol{c}_{\hat{y}_j}\|)}
\end{equation}
\begin{equation}\label{rd}
    \boldsymbol{c}_k = \frac{1}{|\{l:\hat{y}_l=k\}|} \sum_{\hat{y}_l=k} \boldsymbol{h}_l
\end{equation}
where $\hat{y}_l$ is cluster label of node $v_l$, $\boldsymbol{c}_k$ is center embedding of $k$-th cluster, $\sigma$ is an activation function (e.g. Sigmoid) and $\gamma>0$ balances the influence of two types of distances. Smaller $d_i$ and $d_j$ imply more reliable node representation, thus a more reliable difference measure between $\boldsymbol{h}_i$ and $\boldsymbol{h}_j$. $RD_{ij}$ is strictly increasing w.r.t. \(d_{ij}\), and strictly decreasing w.r.t. \(d_i\) and \(d_j\). 
Specifically, for $\gamma > 0$ and a strictly increasing function
$\sigma: \mathbb{R} \to \mathbb{R}^+$, the monotonicity of $RD_{ij}$ follows
directly from its partial derivatives.
By the quotient rule, we have
\begin{equation}
\frac{\partial RD_{ij}}{\partial d_{ij}}
=\frac{\gamma\,\sigma'(\gamma d_{ij})\,\sigma(d_i^c+d_j^c)}
{\sigma(d_i^c+d_j^c)^2}
=\frac{\gamma\,\sigma'(\gamma d_{ij})}
{\sigma(d_i^c+d_j^c)}
\end{equation}
which is strictly positive since $\sigma' > 0$ and $\sigma(\cdot) > 0$.
Similarly, the partial derivative with respect to $d_i^c$ is given by
\begin{equation}
\frac{\partial RD_{ij}}{\partial d_i^c}
=-\frac{\sigma(\gamma d_{ij})\,\sigma'(d_i^c+d_j^c)}
{\sigma(d_i^c+d_j^c)^2}
\end{equation}
which is strictly negative, and the same conclusion holds for $d_j^c$.
Therefore, $RD_{ij}$ increases monotonically with $d_{ij}$ and decreases
monotonically with respect to both $d_i$ and $d_j$.


\subsubsection{Variation}
Representation distortion arise when message passing between dissimilar nodes draws their embeddings closer and away from correct classification regions. Thus, we compute variation of reliable difference after message passing to measure the effect.
With embedding space \textbf{b}efore $l_b$-th and \textbf{a}fter $l_a$-th layer aggregation as $\boldsymbol{H}_b^{l_b}=\sigma(\boldsymbol{H}^{l_b-1}\boldsymbol{W}+\boldsymbol{b})$ and $\boldsymbol{H}_a^{l_a}=\sigma(\boldsymbol{\hat{A}}_{sym}(\boldsymbol{H}^{l_a-1}\boldsymbol{W}+\boldsymbol{b}))$, we compute $RD_{ij}^{l_b,b}$ and $RD_{ij}^{l_a,a}$ for each connected pair $(v_i,v_j)$, and define the variation as below.
\begin{equation}\label{vrd}
    VRD_{ij}^{l_b,l_a}=RD_{ij}^{l_b,b} - RD_{ij}^{l_{a},a},\quad l_a,l_b\in\{1,2,...,N_l\}
\end{equation}
where $N_l$ is the total number of message passing layers and $l_b\le l_a$. In this paper, we set $l_a=l_b=1$, focus on the effect of the first-round message passing. Thus, we abbreviate the notations as $RD_{ij}^b$, $RD_{ij}^a$ and $VRD_{ij}$. In summary, $VRD_{ij}$ measures the decline of reliable difference after one-layer message passing. A higher $VRD_{ij}$ indicates a greater negative effect of message passing between $(v_i,v_j)$.

\subsubsection{Modulation}
While $VRD_{ij}$ tends to increase with higher $RD_{ij}^b$ and lower $RD_{ij}^a$, an extremely small $d_{ij}^a$---indicating that $v_i$ and $v_j$ become highly similar after aggregation---often reflects effective neighbor aggregation in homophilic regions rather than representation distortion. To suppress the benign convergence and prevent overestimation of $VRD_{ij}$ in such case, we introduce a modulation:
\begin{equation}\label{mvrd}
    MVRD_{ij}=\sigma\, (\eta\cdot d_{ij}^a)\cdot VRD_{ij}
\end{equation}
where $\sigma$ denotes activation function (e.g. Sigmoid) and $\eta$ balances the influence of modulation.


\subsection{Active Learning for Edge Selection}
To scale our LM-enhanced message passing for large graphs, it's impractical to enhance all edges with $O(E)$ complexity. Thus, we use an active learning strategy with heuristic MVRD to identify and enhance edges prone to suffer from message passing. 
However, active learning strategy struggle with unstable model weights and node representations in early training stages, leading to suboptimal edge selection. Thus, we introduce a weight-free auxiliary model for stable guidance. Specifically, a weight-free 2-layer GCN can be formulated as $\mathcal{M}^{wf}:\sigma(\boldsymbol{\hat{A}}_{sym}\cdot\sigma(\boldsymbol{\hat{A}}_{sym}\boldsymbol{X}))$, while its paired weight-based one is $\mathcal{M}^{wb}:\sigma(\boldsymbol{\hat{A}}_{sym}\cdot \sigma(\boldsymbol{\hat{A}}_{sym}\boldsymbol{X}\boldsymbol{W}_0)\boldsymbol{W}_1)$. 
We then compute $MVRD_{ij}^{wf}$ and $MVRD_{ij}^{wb}$ as Equation \ref{rd}-\ref{mvrd} and introduce a time-sensitive weight $\lambda$ to fuse them as below:
\begin{equation}\label{time_sensitive}
    \lambda=\omega\cdot cos(\frac{\pi\cdot n_e}{N_e})+\phi, \ \text{with}\ N_e=k\frac{\mathcal{B}}{\mathcal{I}}
\end{equation}
\begin{equation}
    MVRD_{ij}=\lambda\cdot MVRD_{ij}^{wf}+(1-\lambda)\cdot MVRD_{ij}^{wb}
\end{equation}
where $n_e$ is current training epoch, $\mathcal{B}$, $\mathcal{I}$ and $k$ are budget, epoch interval and batch size for query. During training, we select top-$k$ edges with highest $MVRD$ scores every $\mathcal{I}$ epochs to query LM for connection analysis to enhance messages. Training stops at budget exhaustion or patience limit.

\begin{table*}
\centering
\fontsize{8.5pt}{8.5pt}\selectfont
\caption{Categorization of TAG datasets. H-Cat is based on $\mathcal{H}_{node}$ and $\mathcal{H}_{edge}$, while MP-Cat reflects the performance shift after message passing. Specifically, datasets exhibiting performance decline after message passing are classified as malignant, improvements as benign, and others as ambiguous.}
\label{tab:taxonomy}
\begin{tabular}{c|c|c|cc|cc|cc}
\toprule
\textbf{H-Cat.} & \textbf{Datasets} & \textbf{MP-Cat.} & $\mathcal{H}_{node}$ & $\mathcal{H}_{edge}$ & 2-MLP & 4-MLP & 2-GCN & 4-GCN \\
\midrule
\multirow{7.5}{*}{Heterophily} & Cornell & \multirow{6}{*}{Malignant} & 0.1155 & 0.1241 & \textbf{0.8654 ± 0.0674} & \textbf{0.8333 ± 0.0948} & 0.6474 ± 0.0529 & 0.5321 ± 0.1347 \\
& Texas & & 0.0661 & 0.0643 & \textbf{0.8462 ± 0.0000} & \textbf{0.8205 ± 0.0363} & 0.6090 ± 0.0706 & 0.5705 ± 0.0245 \\
& Washington & & 0.1610 & 0.1507 & \textbf{0.8404 ± 0.0662} & \textbf{0.8511 ± 0.0796} & 0.6543 ± 0.0268 & 0.6383 ± 0.0174 \\
& Wisconsin & & 0.1609 & 0.1808 & \textbf{0.8796 ± 0.0685} & \textbf{0.8981 ± 0.0717} & 0.5972 ± 0.1073 & 0.5324 ± 0.1204 \\
& arxiv23 & & 0.2966 & 0.6443 & \textbf{0.7811 ± 0.0035} & \textbf{0.7774 ± 0.0028} & 0.7781 ± 0.0021 & 0.7705 ± 0.0017 \\
& Children & & 0.4559 & 0.4043 & \textbf{0.6199 ± 0.0071} & \textbf{0.6136 ± 0.0064} & 0.6054 ± 0.0085 & 0.5880 ± 0.0192 \\
\cmidrule(lr){2-9}
& Amazon & Benign & 0.3757 & 0.3804 & 0.4275 ± 0.0087 & 0.4346 ± 0.0224 & \textbf{0.4543 ± 0.0118} & \textbf{0.4495 ± 0.0052} \\
\midrule
\multirow{10.5}{*}{Homophily} & Pubmed & \multirow{2}{*}{Malignant} & 0.7924 & 0.8024 & \textbf{0.9471 ± 0.0043} & \textbf{0.9473 ± 0.0036} & 0.9349 ± 0.0029 & 0.9326 ± 0.0011 \\
& History & & 0.7805 & 0.6398 & \textbf{0.8616 ± 0.0052} & \textbf{0.8554 ± 0.0059} & 0.8540 ± 0.0060 & 0.8483 ± 0.0053 \\
\cmidrule(lr){2-9}
& Cora & \multirow{6}{*}{Benign} & 0.8252 & 0.8100 & 0.8034 ± 0.0161 & 0.7947 ± 0.0244 & \textbf{0.8743 ± 0.0190} & \textbf{0.8840 ± 0.0086} \\
& citeseer & & 0.7440 & 0.7841 & 0.7371 ± 0.0116 & 0.7351 ± 0.0095 & \textbf{0.7853 ± 0.0128} & \textbf{0.7857 ± 0.0167} \\
& Photo & & 0.7850 & 0.7351 & 0.7124 ± 0.0006 & 0.7133 ± 0.0020 & \textbf{0.8541 ± 0.0065} & \textbf{0.8577 ± 0.0023} \\
& Computers & & 0.8528 & 0.8228 & 0.6073 ± 0.0044 & 0.6042 ± 0.0016 & \textbf{0.8710 ± 0.0028} & \textbf{0.8806 ± 0.0024} \\
& Fitness & & 0.9000 & 0.8980 & 0.8969 ± 0.0010 & 0.8958 ± 0.0025 & \textbf{0.9277 ± 0.0002} & \textbf{0.9286 ± 0.0004} \\
& products & & 0.7971 & 0.8081 & 0.8519 ± 0.0011 & 0.8499 ± 0.0017 & \textbf{0.8871 ± 0.0006} & \textbf{0.8801 ± 0.0015} \\
\cmidrule(lr){2-9}
& wikics & \multirow{2}{*}{Ambiguous} & 0.6579 & 0.6543 & 0.8597 ± 0.0060 & \textbf{0.8599 ± 0.0046} & \textbf{0.8672 ± 0.0073} & 0.8549 ± 0.0013 \\
& tolokers & & 0.6344 & 0.5945 & \textbf{0.7793 ± 0.0096} & 0.7824 ± 0.0044 & 0.7783 ± 0.0072 & \textbf{0.7848 ± 0.0038} \\
\bottomrule
\end{tabular}
\end{table*}

\section{Experiments} 
\subsection{Experiment Setup}
\subsubsection{Datasets}
Since most heterophilic benchmarks lack raw textual attributes, we collect 16 publicly available text-attributed graph datasets following recent studies \cite{liu2023one, yan2023comprehensive}, covering Academic Webpage, Citation, E-Commerce, Knowledge, and Anomaly domains. 
All graphs are treated as undirected with self-loops removed. We largely follow the original data splits provided in prior work \cite{chen2024exploring, ramsundar2019deep, liu2023one, yan2023comprehensive}, and additionally include a subgraph of ogbn-products for large-scale evaluation. Dataset statistics are summarized in Table~\ref{Dataset Statistics}.
For each dataset, we compute node and edge homophily scores $\mathcal{H}_{node}$ and $\mathcal{H}_{edge}$, and evaluate using both graph-aware (GCN) and graph-agnostic (MLP) models with 2- and 4-layer configurations. Results are summarized in Table \ref{tab:taxonomy}. 
Based on homophily metrics and performance shifts after message passing, datasets are categorized into \emph{benign}, \emph{malignant}, and \emph{ambiguous}, following \cite{luan2024heterophilic}. 
Notably, we observe that some datasets with high homophily scores still exhibit performance degradation after message passing, highlighting that homophily metrics alone are insufficient for identifying challenging heterophilic scenarios.
\begin{itemize} \label{dataset_content}
	\item \textbf{Academic Webpage.} Cornell, Texas, Washington, and Wisconsin consist of university web pages, where nodes are pages, edges are hyperlinks, and node texts are the original webpage contents \cite{yan2023comprehensive}.
	\item \textbf{Citation Networks.} Cora, Citeseer, Pubmed, and arxiv23 are citation graphs. For Cora, Citeseer, and Pubmed, we extract raw texts following \cite{he2023harnessing, chen2024exploring}, while arxiv23 directly adopts paper abstracts from \cite{he2023harnessing}.
	\item \textbf{E-Commerce.} History, Children, Computers, Photo, Fitness, Amazon, and products are derived from Amazon product networks \cite{ni2019justifying, jure2014snap}. Nodes represent items with textual descriptions or reviews, and edges indicate frequent co-purchasing or co-viewing relations.
	\item \textbf{Knowledge.} WikiCS is a Wikipedia hyperlink network, where nodes are pages with entry texts and labels corresponding to categories \cite{mernyei2020wiki}.
	\item \textbf{Anomaly Detection.} Tolokers models worker interaction graphs from the Toloka platform, with node texts describing worker profiles and performance, and labels indicating banned accounts \cite{platonov2023critical, chen2024text}.
\end{itemize}

\begin{table}
    \caption{Statistics of the datasets.}
    \label{Dataset Statistics}
    \centering
    \setlength{\tabcolsep}{4pt}
    \begin{tabular}{ccccccc}
        \toprule
        Datasets & Nodes & Edges & Domains & Class & Split \\
        \midrule
        Cornell & 191 & 274 & Acad Webpage & 5 & 48/32/20 \\
        Texas & 187 & 280 & Acad Webpage & 5 & 48/32/20 \\
        Washington & 229 & 365 & Acad Webpage & 5 & 48/32/20 \\
        Wisconsin & 265 & 459 & Acad Webpage & 5 & 48/32/20 \\
        arxiv23 & 46,198 & 38,863 & CS Citation & 38 & 60/20/20 \\
        Children & 76,875 & 1,162,522 & E-Commerce & 24 & 60/20/20 \\
        Amazon & 24,492 & 93,050 & E-Commerce & 5 & 50/25/25 \\
        Pubmed & 19,717 & 44,324 & Bio Citation & 3 & 60/20/20 \\
        History & 41,551 & 251,590 & E-Commerce & 12 & 60/20/20 \\
        Cora & 2,708 & 5,278 & CS Citation & 7 & 60/20/20 \\
        citeseer & 3,186 & 4,225 & CS Citation & 6 & 60/20/20 \\
        Photo & 48,362 & 436,891 & E-Commerce & 12 & 60/20/20 \\
        Computers & 87,229 & 628,274 & E-Commerce & 10 & 72/17/11 \\
        Fitness & 173,055 & 1,510,067 & E-Commerce & 13 & 20/10/70 \\
        wikics & 11,701 & 215,603 & Knowledge & 10 & 60/20/20 \\
        tolokers & 11,758 & 519,000 & Anomaly & 2 & 50/25/25 \\
        products & 316,513 & 9,668,861 & E-Commerce & 39 & 8/2/90 \\
        \bottomrule
    \end{tabular}
\end{table}

\subsubsection{Baselines} 
We compare our LEMP4HG against four categories of baselines: MLP, classic GNNs (GCN \cite{kipf2016semi}, SAGE \cite{hamilton2017inductive}, GAT \cite{velivckovic2017graph}, RevGAT \cite{li2021training}, GCN-Cheby \cite{defferrard2016convolutional}, JKNet \cite{xu2018representation}, APPNP \cite{gasteiger2018predict}), heterophily-specific GNNs (H2GCN \cite{zhu2020beyond}, GCNII \cite{chen2020simple}, FAGCN \cite{bo2021beyond}, GPRGNN \cite{chien2020adaptive}, JacobiConv \cite{wang2022powerful}, GBK-GNN \cite{du2022gbk}, OGNN \cite{du2022gbk}, SEGSL \cite{zou2023se}, DisamGCL \cite{zhao2024disambiguated}, GNN-SATA \cite{yang2024graph}), LM-enhanced GNNs (SAGE- and RevGAT-backboned TAPE \cite{he2023harnessing}, LLM4HeG \cite{wu-etal-2025-exploring}).
\subsubsection{Experiment Platform}
All experiments are conducted on a single NVIDIA A100 GPU with 80GB memory. The software environment includes PyTorch 2.4.1 with CUDA 12.0 support and PyTorch Geometric 2.6.1. 

\subsubsection{Implementation}
We adopt Qwen-turbo as LM to generate connection analysis via API calls and DeBERTa-base \cite{he2021deberta} as SLM to encode texts. Following the common practice, we randomly split nodes into train, validation and test sets as Table \ref{Dataset Statistics}, where all experiments are performed with 4 runs and reported as average results.

\subsection{Hyper-parameter Settings}
\subsubsection{SLM Finetuning} We employ DeBERTa-base \cite{he2021deberta} as our SLM for text encoding. The model is finetuned for semi-supervised node classification by appending a one-layer MLP classification head, trained with a cross entropy loss function incorporating label smoothing (0.3). During finetuning, we adopt a training schedule of 4 or 8 epochs, with an initial warm-up phase of 0.6 epochs to stabilize optimization. The learning rate is set to 2e-5, accompanied by a weight decay factor of 0.0 to prevent over-regularization. Dropout regularization is applied with a rate of 0.3 on fully connected layers, and an attention dropout rate of 0.1 is used to mitigate overfitting within the self-attention mechanism. Gradient accumulation steps are set to 1, and training batches consist of 9 samples per device. The parameter settings and training protocol are largely aligned with TAPE \cite{he2023harnessing}.


\subsubsection{GNN Training}We employ a two-layer GCN as the backbone of LEMP4HG, with hidden representations of 128 dimensions. The model is trained for a maximum of 500 epochs using early stopping with a patience of 50 epochs. Optimization is performed using a learning rate of 2e-2, weight decay of 5e-4, and a dropout rate of 0.5. For message synthesis, we adopt the Sigmoid activation function as $\sigma$ in Equation \ref{concat}, and set the gating coefficient to $\beta = 0.5$ in Equation \ref{gate}.

\subsubsection{Heuristic Definition} For reliable difference in Equation \ref{rd}, we set $\gamma = 1.0$ and adopt the Sigmoid function as $\sigma$. For variation in Equation \ref{vrd}, we set $l_a=l_b=1$. For modulation in Equation \ref{mvrd}, we set $\eta = 0.8$ and adopt the Sigmoid function as $\sigma$. Additionally, all distance calculations are batch-normalized, and the final MVRD scores are weighted by $\hat{A}_{\text{sym}}$ to ensure consistency with the message-passing mechanism. The sensitivity analysis of hyper-parameters in the MVRD heuristic definition is provided in the Section \ref{hyper-parameter-study}

\subsubsection{Active Leaning for Edge Selection} For time-sensitive weight in Equation \ref{time_sensitive}, we set $\omega=0.5$, $\phi=0.5$, $\mathcal{I}=10 \text{ epochs}$. Additionally, the high-dimensional input $X$ from SLM's hidden layer (e.g. 768-dimension) may hinder the effectiveness of $\mathcal{M}^{wf}$ without a projection matrix. Thus, we apply PCA (e.g. reducing to 128 dimensions) before feeding $X$ into $\mathcal{M}^{wf}$.

\begin{table*}
\centering
\fontsize{8.5pt}{9.5pt}\selectfont
\setlength{\tabcolsep}{0.9pt}
\caption{Evaluation of our LEMP4HG and baselines on various text-attributed graphs. "OOT" and "OOM" denote runtime or memory limits failures. "+T" denotes enhanced by TAPE. \textbf{Bold} numbers indicate the top-3 performances and rankings.}
\label{main_results}
\begin{tabular}{c|ccccccc|ccccccccc|cc}
\toprule
\multirow{2.5}{*}{Models} & \multicolumn{7}{c|}{Heterophilic Graph} & \multicolumn{9}{c|}{Homophilic Graph} & \multicolumn{2}{c}{Rank}\\
\cmidrule(lr){2-8} \cmidrule(lr){9-17} \cmidrule(lr){18-19}
 & Cornell & Texas & Wash. & Wis. & arxiv23 & Child & Amazon & Pubmed & History & Cora & citeseer & Photo & Comp. & Fitness & wikics & tolokers & w/ $\text{f}_4$ & w/o $\text{f}_4$ \\
\midrule
MLP & \textbf{0.8654} & 0.8462 & 0.8404 & 0.8796 & 0.7811 & 0.6199 & 0.4275 & 0.9471 & 0.8616 & 0.8034 & 0.7371 & 0.7121 & 0.6065 & 0.8969 & 0.8597 & 0.7793 & 13.69 & 15.75 \\
\midrule
GCN & 0.6346 & 0.6026 & 0.6596 & 0.5972 & 0.7785 & 0.6083 & 0.4558 & 0.9354 & 0.8559 & 0.8762 & 0.7853 & \textbf{0.8564} & 0.8735 & \textbf{0.9282} & 0.8700 & 0.7820 & 12.44 & 9.25 \\
SAGE & 0.8269 & 0.8269 & 0.8564 & 0.8935 & 0.7861 & 0.6245 & \textbf{0.4648} & 0.9475 & 0.8649 & 0.8531 & 0.7813 & 0.8518 & 0.8727 & 0.9240 & 0.8771 & 0.7885 & 7.13 & 6.67 \\
GAT & 0.4808 & 0.5962 & 0.5532 & 0.4769 & 0.7622 & 0.5824 & 0.4520 & 0.8875 & 0.8441 & 0.8725 & 0.7841 & 0.8545 & \textbf{0.8738} & 0.9261 & 0.8533 & 0.7821 & 14.94 & 12.25 \\
RevGAT & 0.8397 & 0.8205 & \textbf{0.8777} & 0.8935 & 0.7798 & 0.6195 & 0.4590 & \textbf{0.9484} & 0.8645 & 0.8085 & 0.7551 & 0.7839 & 0.7597 & 0.9083 & 0.8665 & \textbf{0.7968} & 10.06 & 11.00 \\
Cheby & \textbf{0.8654} & 0.8462 & 0.8404 & 0.8796 & 0.7811 & 0.6199 & 0.4275 & 0.9471 & 0.8616 & 0.8034 & 0.7371 & 0.7114 & 0.6045 & 0.8969 & 0.8597 & 0.7793 & 13.81 & 15.92 \\
JKNet & 0.6603 & 0.6410 & 0.7181 & 0.6667 & 0.7774 & 0.6031 & 0.4551 & 0.9314 & 0.8537 & \textbf{0.8821} & 0.7845 & 0.8545 & \textbf{0.8739} & \textbf{0.9282} & 0.8629 & 0.7838 & 12.38 & 9.75 \\
APPNP & 0.6474 & 0.6538 & 0.7500 & 0.6250 & 0.7762 & 0.6241 & 0.4534 & 0.9066 & 0.8569 & \textbf{0.8821} & \textbf{0.7927} & 0.8446 & 0.8647 & 0.9279 & 0.8754 & 0.7809 & 12.63 & 10.08 \\
\midrule
H2GCN & 0.6795 & 0.7244 & 0.8138 & 0.7639 & 0.7761 & 0.6126 & 0.4071 & 0.9473 & 0.8383 & 0.8324 & 0.7712 & 0.8441 & 0.8632 & 0.9178 & 0.8660 & 0.7815 & 15.00 & 13.92 \\
GCNII & 0.8013 & 0.8013 & 0.8564 & 0.8750 & 0.7832 & 0.6223 & 0.4429 & 0.9483 & 0.8630 & 0.8352 & 0.7441 & 0.8493 & \textbf{0.8736} & 0.9137 & 0.8674 & 0.7861 & 10.13 & 9.25 \\
FAGCN & 0.7051 & 0.7949 & 0.7074 & 0.8102 & 0.7446 & 0.6287 & 0.4319 & 0.8859 & 0.7784 & 0.8191 & 0.7555 & 0.8080 & 0.7216 & 0.7790 & 0.8655 & 0.7812 & 16.50 & 15.83 \\
GPR & 0.8269 & 0.8333 & 0.8564 & 0.8796 & 0.7815 & \textbf{0.6316} & 0.4554 & 0.9470 & 0.8583 & 0.8794 & 0.7861 & 0.8467 & 0.8728 & 0.9277 & 0.8755 & 0.7813 & 8.56 & 8.25 \\
Jacobi & 0.7756 & 0.7692 & 0.7766 & 0.8426 & 0.7153 & 0.5981 & 0.4554 & 0.9473 & 0.8543 & 0.8734 & 0.7810 & 0.8432 & 0.8610 & 0.9238 & 0.8778 & 0.7817 & 13.06 & 11.58 \\
GBK & 0.8333 & 0.8397 & 0.8085 & 0.8889 & 0.7617 & 0.4961 & 0.4274 & 0.9476 & 0.8403 & 0.8250 & 0.7649 & 0.7659 & 0.6954 & 0.8771 & 0.8723 & 0.7799 & 14.69 & 15.92 \\
OGNN & 0.8462 & 0.8397 & 0.8564 & \textbf{0.8981} & 0.7820 & 0.6250 & 0.4366 & \textbf{0.9487} & 0.8633 & 0.8066 & 0.7504 & 0.7914 & 0.7693 & 0.9095 & 0.8684 & 0.7803 & 10.38 & 11.83 \\
SEGSL & 0.8333 & \textbf{0.8590} & 0.8564 & \textbf{0.9028} & OOT & OOT & OOT & OOT & OOT & 0.8191 & 0.7680 & OOT & OOT & OOT & OOT & OOT & 8.00 & 14.50 \\
Disam & 0.8462 & 0.8141 & 0.8404 & 0.8704 & 0.7801 & OOM & 0.4410 & 0.9476 & 0.8604 & 0.8103 & 0.7343 & OOM & OOM & OOM & 0.8651 & 0.7835 & 13.08 & 13.50 \\
SATA & 0.8141 & 0.8077 & 0.8457 & 0.8935 & OOM & OOM & 0.4237 & 0.9453 & OOM & 0.8043 & 0.7339 & OOM & OOM & OOM & 0.8602 & 0.7815 & 15.70 & 18.50 \\
\midrule
SAGE+T & \textbf{0.8718} & 0.8526 & 0.8670 & 0.8889 & \textbf{0.8023} & \textbf{0.6316} & 0.4639 & 0.9480 & \textbf{0.8677} & 0.8771 & 0.7837 & \textbf{0.8587} & 0.8733 & \textbf{0.9315} & \textbf{0.8823} & 0.7848 & \textbf{4.06} & \textbf{4.19} \\
RevGAT+T & \textbf{0.8846} & \textbf{0.8590} & \textbf{0.8777} & \textbf{0.9074} & \textbf{0.7995} & \textbf{0.6285} & \textbf{0.4722} & 0.9480 & \textbf{0.8664} & 0.8439 & 0.7774 & 0.8002 & 0.7640 & 0.9215 & \textbf{0.8824} & \textbf{0.7991} & \textbf{5.25} & \textbf{5.52} \\
LLM4HeG & 0.8524 & 0.8362 & 0.8245 & 0.8696 & - & - & - & 0.9440 & - & 0.8283 & 0.7418 & - & - & - & - & - & 11.86 & 13.12 \\
\midrule
LEMP & 0.8526 & 0.8269 & 0.8564 & \textbf{0.8981} & 0.7853 & 0.6137 & 0.4590 & \textbf{0.9485} & 0.8590 & 0.8803 & \textbf{0.7888} & \textbf{0.8564} & 0.8735 & \textbf{0.9282} & 0.8729 & 0.7867 & 5.63 & 5.33 \\
LEMP+T & \textbf{0.8654} & \textbf{0.8590} & \textbf{0.8777} & 0.8565 & \textbf{0.8003} & 0.6179 & \textbf{0.4675} & \textbf{0.9484} & \textbf{0.8662} & \textbf{0.8826} & \textbf{0.7943} & \textbf{0.8591} & 0.8729 & \textbf{0.9303} & \textbf{0.8825} & \textbf{0.7897} & \textbf{3.69} & \textbf{3.73} \\
\bottomrule
\end{tabular}
\end{table*}

\subsection{Main Results}
We evaluate LEMP4HG on 16 text-attributed graph datasets and compare it with four categories of baselines (Table~\ref{main_results}). Compared with its GCN backbone, LEMP4HG achieves consistent improvements on 13 out of 16 datasets, demonstrating the effectiveness of LM-enhanced message passing. The remaining datasets (Photo, Computers, Fitness) are large-scale and highly homophilic, where conventional message passing is already sufficient and additional LM-enhanced messages may introduce noise.

To ensure fair comparison, we report average rankings both with and without $\text{f}_4$, which contains four small heterophilic datasets with high variance. Under both settings, LEMP4HG consistently outperforms MLP, classical GNNs, and heterophily-specific models, indicating strong robustness across diverse graph structures.

Although TAPE-enhanced methods (TAPE+S and TAPE+R) achieve slightly better average rankings, their gains mainly rely on ensembling multiple independently trained models, whereas LEMP4HG operates as a single-model approach. Moreover, integrating TAPE into our framework (LEMP+T) further improves performance, suggesting that LEMP4HG is complementary to TAPE-based enhancements.

\begin{figure}
    \centering
    \scalebox{0.48}{
    \includegraphics[width=\textwidth]{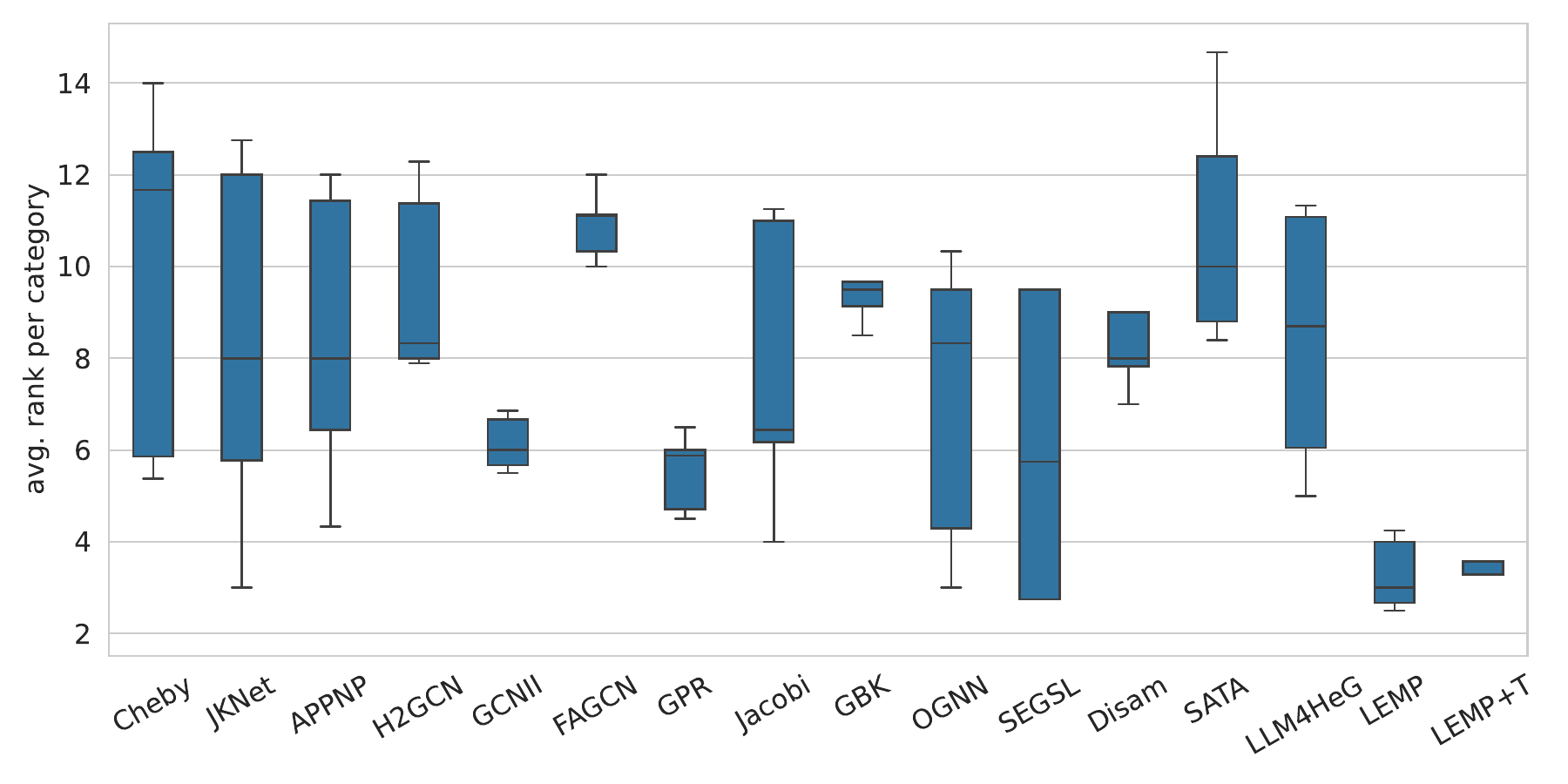}
    }
    \captionof{figure}{Rank distribution on 5 dataset categories. Lower the box, more robust the model.}
    \label{rank_box}
\end{figure}

\textbf{Fair Evaluation.}
We further evaluate robustness by averaging model rankings over five dataset categories (heterophilic, homophilic, malignant, benign, ambiguous). As shown in Table~\ref{robust_category} and Fig.~\ref{rank_box}, most heterophily-specific methods fail to generalize across categories, while LEMP4HG consistently maintains strong and stable performance.

\begin{table*}[htbp]
\centering
\setlength{\tabcolsep}{2.5pt}
\caption{Average ranking of models across dataset categories. \textbf{Bold} numbers indicate the top-3 rankings.}
\label{robust_category}
\begin{tabular}{ccccccccccccccccc}
\toprule
Dataset & Cheby & JKNet & APPNP & H2GCN & GCNII & FAGCN & GPR & Jacobi & GBK & OGNN & SEGSL & Disam & SATA & LLM4HeG & LEMP & LEMP+T \\
\midrule
heterophilic & 5.86 & 12.00 & 11.43 & 12.29 & 6.86 & 11.14 & 4.71 & 11.00 & 9.14 & 4.29 & \textbf{2.75} & 7.83 & 8.80 & 5.00 & \textbf{4.00} & \textbf{3.29} \\
homophilic & 11.67 & 5.78 & 6.44 & 7.89 & \textbf{5.67} & 11.11 & 6.00 & 6.44 & 9.67 & 8.33 & 9.50 & 9.00 & 12.40 & 11.33 & \textbf{3.00} & \textbf{3.57} \\
malignant & 5.38 & 12.75 & 12.00 & 11.38 & 6.00 & 12.00 & 5.88 & 11.25 & 8.50 & \textbf{3.00} & \textbf{2.75} & 7.00 & 8.40 & 6.40 & 4.25 & \textbf{3.57} \\
benign & 12.50 & \textbf{3.00} & 4.33 & 8.33 & 6.67 & 10.33 & 4.50 & 6.17 & 10.67 & 10.33 & 9.50 & 12.00 & 14.67 & 11.00 & \textbf{2.67} & \textbf{3.57} \\
ambiguous & 14.00 & 8.00 & 8.00 & 8.00 & \textbf{5.50} & 10.00 & 6.50 & \textbf{4.00} & 9.50 & 9.50 & 5.75 & 8.00 & 10.00 & - & \textbf{2.50} & \textbf{2.14} \\

\bottomrule
\end{tabular}
\end{table*}

\subsection{Ablation Study}
We conduct ablation studies on heuristic design and message synthesis (Table~\ref{ablation}). For heuristics, we compare MVRD, VRD (without modulation), and FeatDiff. Note that all edges in the small $\text{f}_4$ datasets are enhanced by LM, resulting in identical behavior across heuristics.
For message synthesis, LEMP4HG fuses preliminary messages with paired node textual embeddings via a gating mechanism. We evaluate three variants: w/o m (no preliminary messages), w/o n (no node textual embeddings), and w/o mn (GCN backbone). Results averaged over four runs demonstrate that both heuristic modulation and message synthesis are essential.
Specifically, VRD introduces noisy enhancements by misattributing benign convergence to representation distortion, while FeatDiff fails to accurately identify node pairs most affected by message passing. Removing preliminary messages (w/o m) limits semantic information, whereas removing textual embeddings (w/o n) degrades multi-hop alignment during aggregation.

\begin{table*}
\centering
\fontsize{8.5pt}{11.0pt}\selectfont
\caption{Ablation studies: heuristic definition and message synthesis. \textbf{Bold} indicates the optimal performance, while underlined ones the runner-up. "\textbackslash" indicates consistency with no ablation.}
\label{ablation}
\setlength{\tabcolsep}{3.5pt}
\begin{tabular}{c|c|cccccccccccccc}
\toprule
\textbf{Ablation} & \textbf{Varient} & Cornell & Texas & Wash. & Wis. & arxiv23 & Child & Amazon & Pubmed & History & Cora & citeseer & wikics & tolokers \\
\midrule
\multirow{3}{*}{\shortstack{Heuristic\\Definition}} 
& \textbf{MVRD}          & 0.8526 & 0.8269 & 0.8564 & 0.8981 & \textbf{0.7853} & \textbf{0.6160} & \textbf{0.4590} & \textbf{0.9485} & \textbf{0.8599} & \uline{0.8803} & \uline{0.7888} & \textbf{0.8768} & \uline{0.7867} \\
& VRD           & \textbackslash & \textbackslash & \textbackslash & \textbackslash & 0.7811 & \uline{0.6116} & \uline{0.4578} & \uline{0.9469} & \uline{0.8579} & \textbf{0.8821} & 0.7861 & \uline{0.8721} & 0.7861 \\
& featDiff    & \textbackslash & \textbackslash & \textbackslash & \textbackslash & \uline{0.7829} & 0.6113 & 0.4577 & 0.9453 & 0.8576 & 0.8752 & \textbf{0.7896} & 0.8680 & \textbf{0.7876} \\
\midrule
\multirow{4}{*}{\shortstack{Message\\Synthesis}} 
& \textbf{w/ mn}         & \textbf{0.8526} & \textbf{0.8269} & \textbf{0.8564} & \textbf{0.8981} & \textbf{0.7853} & \uline{0.6160} & \uline{0.4590} & \textbf{0.9485} & \uline{0.8599} & \uline{0.8803} & \textbf{0.7888} & \textbf{0.8768} & \textbf{0.7867} \\
& w/o m         & 0.7628 & \uline{0.7756} & \uline{0.7713} & \uline{0.8519} & \uline{0.7852} & \textbf{0.6237} & 0.4566 & 0.9471 & \textbf{0.8626} & 0.8439 & 0.7633 & 0.8725 & 0.7843 \\
& w/o n         & \uline{0.7756} & 0.8013 & 0.7394 & 0.8472 & 0.7814 & 0.6119 & \textbf{0.4602} & \uline{0.9472} & 0.8582 & \textbf{0.8821} & 0.7880 & \uline{0.8731} & \uline{0.7861} \\
& w/o mn        & 0.5577 & 0.6026 & 0.7074 & 0.5972 & 0.7777 & 0.6094 & 0.4502 & 0.9364 & 0.8509 & 0.8435 & \textbf{0.7888} & 0.8694 & 0.7809 \\
\bottomrule
\end{tabular}
\end{table*}

\subsection{Scalability}
\subsubsection{Budget Allocation}
\begin{figure*}[htbp]
    \centering
    \begin{subfigure}[t]{\textwidth}
        \centering
        \includegraphics[width=\textwidth]{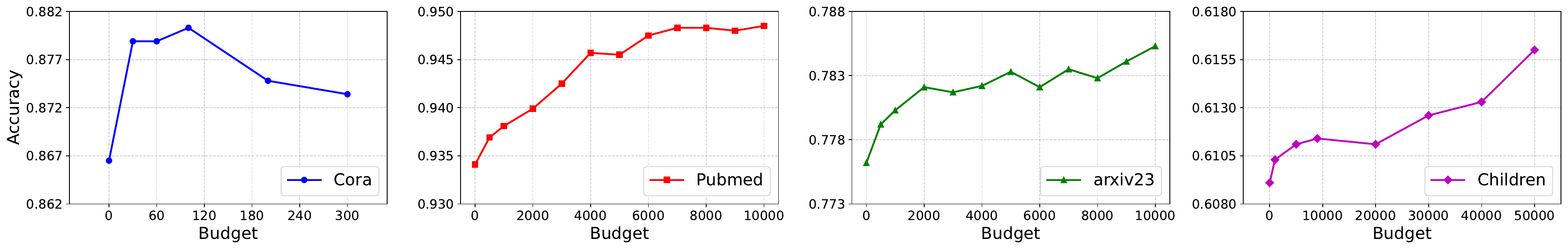}
        \label{fig:ea4_1}
    \end{subfigure}


    \begin{subfigure}[t]{\textwidth}
        \centering
        \includegraphics[width=\textwidth]{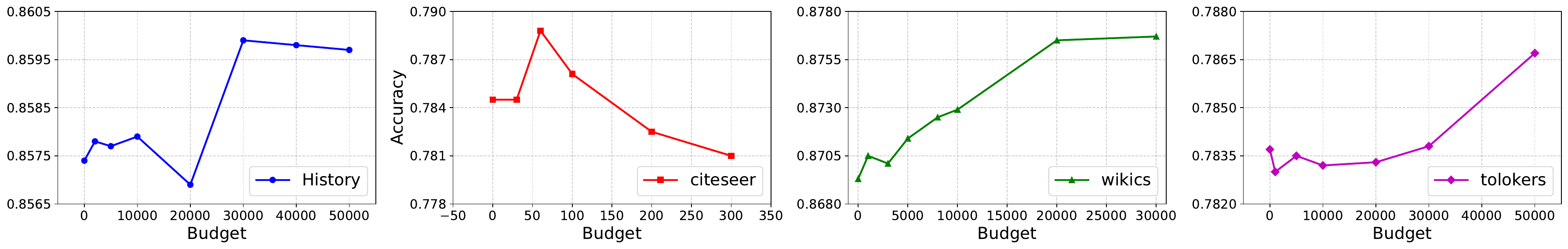}
        \label{fig:ea4_2}
    \end{subfigure}


    \begin{subfigure}[t]{\textwidth}
        \centering
        \includegraphics[width=\textwidth]{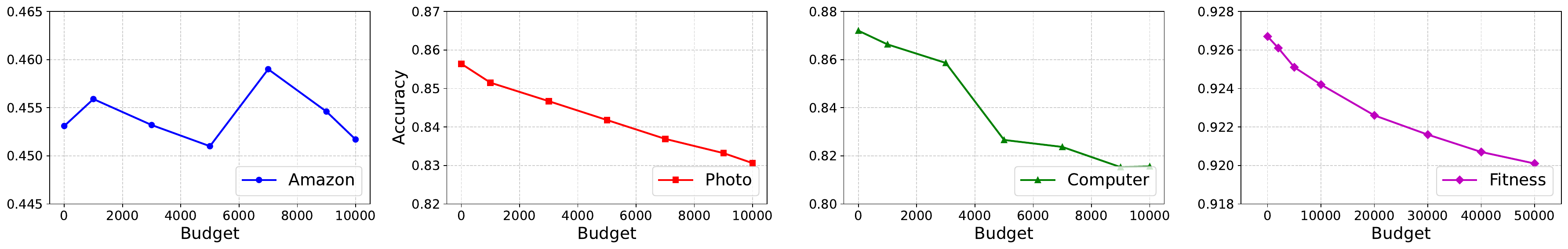}
        \label{fig:ea4_3}
    \end{subfigure}

    \caption{Scalability study across multiple datasets: accuracy v.s. budget under different budgets and domains.}
    \label{fig:ea12}
\end{figure*}

We vary the LM query budget $\mathcal{B}$ to study scalability. Results on 12 datasets are shown in Fig.~\ref{fig:ea12}, while the four small $\text{f}_4$ datasets use full-edge enhancement.
Most datasets exhibit performance improvements as $\mathcal{B}$ increases, with diminishing returns. In contrast, highly homophilic and message-passing-benign datasets (Photo, Computers, Fitness) show performance degradation, as excessive LM-enhanced messages may introduce noise.

\textbf{Budget Allocation Guidelines.\label{budget_guideline}\label{budget}}
Based on empirical observations, we recommend: (1) for homophilic graphs with benign message passing, allocate $\mathcal{B}^*\!\approx\!5\%\cdot|\mathcal{E}|$ for small graphs and $\mathcal{B}^*\!=\!0$ for large graphs; (2) for heterophilic or malignant/ambiguous graphs, progressively increase $\mathcal{B}$ until performance saturates. This strategy consistently improves performance, except for Amazon, where low-quality node texts limit LM effectiveness.

\subsubsection{Language Model Capacity}
We evaluate LEMP4HG using three LMs with increasing capacity (\textit{qwen-turbo}, \textit{qwen-plus}, \textit{qwen-max}) under a fixed budget. Table~\ref{lm_output} shows that LEMP4HG achieves consistent performance across variants.

This robustness stems from encoding LM outputs into message representations and integrating them with node features, which attenuates output variability. Upgrading to more powerful LMs yields marginal gains, indicating that lightweight and cost-efficient models are sufficient for practical deployment.

\begin{table}
    \caption{Performance comparison with different Qwen LM variants. Model capability and API price rank as: qwen-max $>$ qwen-plus $>$ qwen-turbo, while response speed ranks in the reverse order.}
    \label{lm_output}
    \centering
    \begin{tabular}{cccc}
        \toprule
        Dataset & qwen-turbo & qwen-plus & qwen-max \\
        \midrule
        Cora & 0.8803 ± 0.0145 & 0.8812 ± 0.0164 & 0.8807 ± 0.0146 \\
        citeseer & 0.7888 ± 0.0078 & 0.7898 ± 0.0122 & 0.7883 ± 0.0115 \\
        Pubmed & 0.9485 ± 0.0042 & 0.9490 ± 0.0045 & 0.9497 ± 0.0046 \\
        arxiv23 & 0.7853 ± 0.0026 & 0.7860 ± 0.0022 & 0.7868 ± 0.0031 \\
        \bottomrule
    \end{tabular}
\end{table}

\subsubsection{Graph Dataset}
We evaluate scalability on the \textit{products} dataset, a large subgraph of \textit{ogbn-products}. Based on homophily metrics and message-passing effects (Table~\ref{tab:taxonomy}), it is categorized as homophilic and MP-benign.
Following our budget guidelines, such datasets do not require substantial enhancement. Yet, increasing the LM budget still yields steady improvements (Table~\ref{budget_products}), demonstrating scalability without instability.
Table~\ref{acc_products} shows that LEMP4HG outperforms most general-purpose GNNs and several heterophily-specific methods. Although GPRGNN and JacobiConv achieve slightly better accuracy, this aligns with our guideline that lightweight GNNs are already effective for large homophilic graphs.

\begin{table}[t]
	\caption{Scalability study on dataset products: accuracy vs. budget.}
	\label{budget_products}
	\centering
	\small
	\setlength{\tabcolsep}{8pt}
	\begin{tabular}{ccccc}
		\toprule
		Budget & 0 & 10000 & 30000 & 50000 \\
		Accuracy & 0.8882 & 0.8887 & 0.8885 & 0.8901 \\
		\midrule
		Budget & 70000 & 100000 & 150000 & 200000 \\
		Accuracy & 0.8905 & 0.8913 & 0.8927 & 0.8939 \\
		\bottomrule
	\end{tabular}
\end{table}

\begin{table}[t]
	\caption{Model evaluation on dataset Products. 
		Models are grouped by methodology. OOM/OOT results are omitted. 
		Best and second-best results are highlighted in bold and underline.}
	\label{acc_products}
	\centering
	\small
	\renewcommand{\arraystretch}{1.18}
	
	\begin{tabular}{p{2.2cm} p{2.6cm} >{\centering\arraybackslash}p{1.6cm}}
		\toprule
		\textbf{Category} & \textbf{Model} & \textbf{Accuracy} \\
		\midrule
		\multirow{1}{*}{Non-Graph}
		& MLP        & 0.8634 \\
		\midrule
		\multirow{8}{*}{Classical GNNs}
		& GCN        & 0.8868 \\
		& SAGE       & 0.8933 \\
		& GAT        & 0.8829 \\
		& RevGAT     & 0.8859 \\
		& Cheby      & 0.8634 \\
		& JKNet      & 0.8862 \\
		& GCNII      & 0.8738 \\
		& OGNN       & 0.8908 \\
		\midrule
		\multirow{3}{*}{Propagation-based}
		& APPNP      & 0.8634 \\
		& GPR        & \textbf{0.8953} \\
		& Jacobi     & \textbf{0.8968} \\
		\midrule
		\multirow{1}{*}{Ours}
		& LEMP4HG    & \textbf{0.8939} \\
		\bottomrule
	\end{tabular}
\end{table}

\subsection{Hyper-parameter Sensitivity Study} \label{hyper-parameter-study}
We examine the sensitivity of LEMP4HG to its main hyper-parameters across multiple datasets. Unlike some GNNs that rely on extensive hyper-parameter tuning, LEMP4HG, built on a simple 2-layer GCN backbone, performs well with default settings. 
The key tunable parameters are $\gamma$ (scaling pairwise embedding distances) and $\eta$ (modulating benign convergence in MVRD). 
Tables \ref{tab:gamma_eta} report test accuracy on four representative datasets (Pubmed, arxiv23, wikics, and Children) across a grid of $\gamma$ and $\eta$ settings under specific LM query budgets. From these results, we make the following observations:
\begin{itemize}
\item Pubmed and arxiv23 (budget = 10,000): Accuracy is highly stable across all $\gamma$ and $\eta$ values, with variation within ± 0.0002, suggesting MVRD is robust in moderately sized scenarios.
\item wikics and Children (larger graphs / higher budgets): Accuracy varies slightly more, but peaks consistently around $\gamma$ = 1.0 and $\eta$ $\approx$ 0.8, indicating these defaults provide strong generalization.
\end{itemize}
These results suggest that while MVRD is more sensitive in resource-intensive settings, our default configuration ($\gamma$ = 1.0, $\eta$ = 0.8) performs robustly across diverse datasets. 
Other auxiliary parameters, e.g., query interval, are fixed across all experiments.

\begin{table*}[htbp]
\centering
\footnotesize
\setlength{\tabcolsep}{4.5pt}
\caption{Test accuracy across different $\gamma$ and $\eta$ configurations under different datasets and budgets.}
\label{tab:gamma_eta}

\begin{subtable}[t]{0.48\textwidth}
\centering
\caption{Pubmed (budget-10,000)}
\begin{tabular}{c|cccccc|c}
\toprule
$\gamma$ \textbackslash $\eta$ & 0.4 & 0.6 & 0.8 & 1.0 & 1.2 & 1.4 & avg \\
\midrule
0.5  & 0.9473 & 0.9463 & 0.9464 & 0.9468 & 0.9470 & 0.9457 & 0.9466 \\
0.75 & 0.9464 & 0.9466 & 0.9466 & 0.9469 & 0.9470 & 0.9464 & 0.9467 \\
1.0  & 0.9473 & 0.9471 & 0.9462 & 0.9464 & 0.9466 & 0.9468 & 0.9467 \\
1.5  & 0.9466 & 0.9466 & 0.9461 & 0.9473 & 0.9462 & 0.9471 & 0.9467 \\
2.0  & 0.9464 & 0.9471 & 0.9468 & 0.9461 & 0.9471 & 0.9477 & 0.9469 \\
\midrule
avg  & 0.9468 & 0.9467 & 0.9464 & 0.9467 & 0.9468 & 0.9467 & \\
\bottomrule
\end{tabular}
\end{subtable}
\hfill
\begin{subtable}[t]{0.48\textwidth}
\centering
\caption{arxiv23 (budget-10,000)}
\begin{tabular}{c|cccccc|c}
\toprule
$\gamma$ \textbackslash $\eta$ & 0.4 & 0.6 & 0.8 & 1.0 & 1.2 & 1.4 & avg \\
\midrule
0.5  & 0.7833 & 0.7836 & 0.7843 & 0.7846 & 0.7839 & 0.7833 & 0.7838 \\
0.75 & 0.7842 & 0.7830 & 0.7841 & 0.7844 & 0.7839 & 0.7836 & 0.7839 \\
1.0  & 0.7836 & 0.7849 & 0.7843 & 0.7840 & 0.7837 & 0.7834 & 0.7840 \\
1.5  & 0.7841 & 0.7837 & 0.7837 & 0.7837 & 0.7835 & 0.7850 & 0.7840 \\
2.0  & 0.7843 & 0.7841 & 0.7830 & 0.7836 & 0.7840 & 0.7835 & 0.7838 \\
\midrule
avg  & 0.7839 & 0.7839 & 0.7839 & 0.7841 & 0.7838 & 0.7838 & \\
\bottomrule
\end{tabular}
\end{subtable}

\vspace{0.8em}

\begin{subtable}[t]{0.48\textwidth}
\centering
\caption{WikiCS (budget-20,000)}
\begin{tabular}{c|cccccc|c}
\toprule
$\gamma$ \textbackslash $\eta$ & 0.4 & 0.6 & 0.8 & 1.0 & 1.2 & 1.4 & avg \\
\midrule
0.5  & 0.8732 & 0.8743 & 0.8743 & 0.8727 & 0.8737 & 0.8744 & 0.8738 \\
0.75 & 0.8745 & 0.8739 & 0.8738 & 0.8749 & 0.8742 & 0.8740 & 0.8742 \\
1.0  & 0.8725 & 0.8742 & 0.8714 & 0.8749 & 0.8747 & 0.8751 & 0.8738 \\
1.5  & 0.8745 & 0.8738 & 0.8737 & 0.8743 & 0.8749 & 0.8739 & 0.8742 \\
2.0  & 0.8728 & 0.8741 & 0.8720 & 0.8742 & 0.8742 & 0.8738 & 0.8735 \\
\midrule
avg  & 0.8735 & 0.8741 & 0.8730 & 0.8742 & 0.8743 & 0.8742 & \\
\bottomrule
\end{tabular}
\end{subtable}
\hfill
\begin{subtable}[t]{0.48\textwidth}
\centering
\caption{Children (budget-50,000)}
\begin{tabular}{c|cccccc|c}
\toprule
$\gamma$ \textbackslash $\eta$ & 0.4 & 0.6 & 0.8 & 1.0 & 1.2 & 1.4 & avg \\
\midrule
0.5  & 0.6142 & 0.6158 & 0.6166 & 0.6160 & 0.6148 & 0.6144 & 0.6153 \\
0.75 & 0.6155 & 0.6156 & 0.6171 & 0.6154 & 0.6153 & 0.6151 & 0.6157 \\
1.0  & 0.6173 & 0.6167 & 0.6166 & 0.6170 & 0.6142 & 0.6166 & 0.6164 \\
1.5  & 0.6141 & 0.6159 & 0.6129 & 0.6159 & 0.6152 & 0.6163 & 0.6151 \\
2.0  & 0.6148 & 0.6149 & 0.6150 & 0.6132 & 0.6162 & 0.6158 & 0.6150 \\
\midrule
avg  & 0.6152 & 0.6158 & 0.6156 & 0.6155 & 0.6151 & 0.6156 & \\
\bottomrule
\end{tabular}
\end{subtable} 
\end{table*}

\subsection{Case Study}
Both Pubmed and Cornell show malignant outcomes after message passing, with Pubmed displaying homophily while Cornell heterophily. We conduct the case study in the following two aspects.
\subsubsection{MVRD-guided Edge Selection}
\begin{figure*}
    \centering
    \begin{minipage}{0.73\textwidth}
        \centering
        \includegraphics[width=\textwidth]{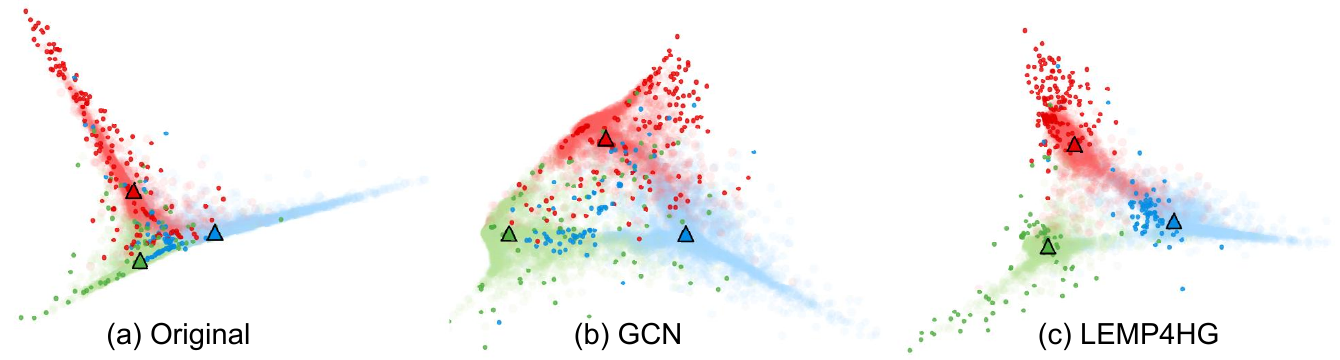}
    \end{minipage}
    \hfill
    \begin{minipage}{0.25\textwidth}
        \centering
        \includegraphics[width=\textwidth]{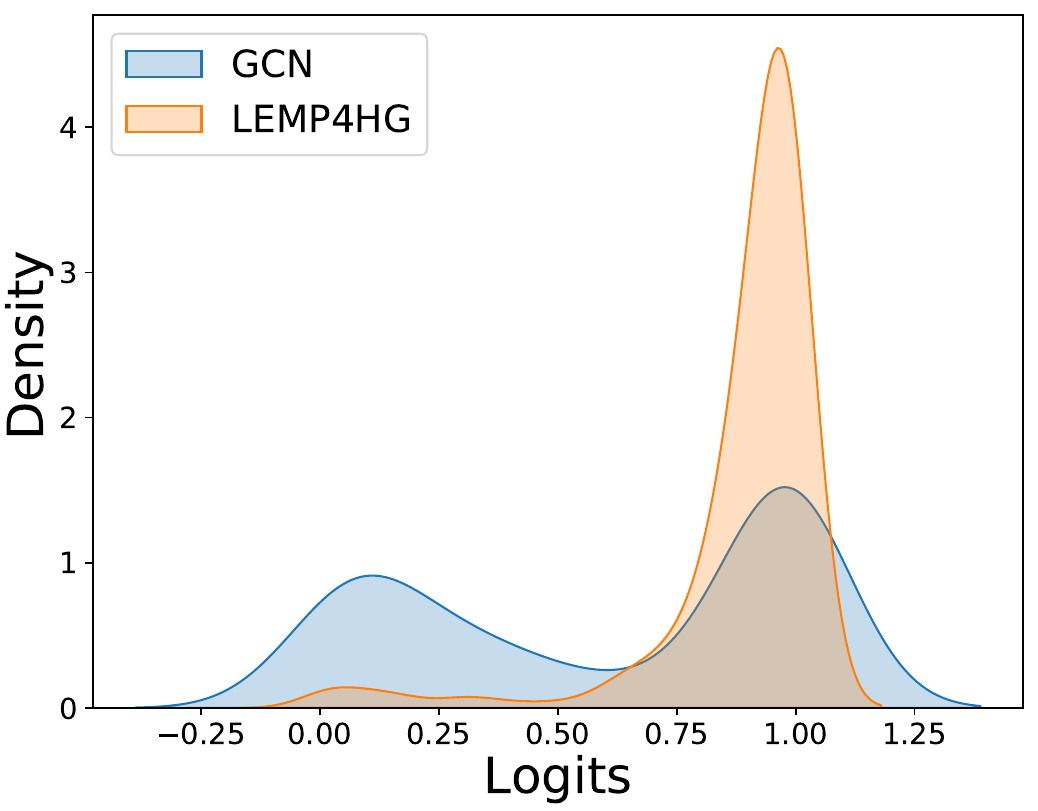}
    \end{minipage}
    \caption{(left) Embedding space before and after message passing. (right) Logits distribution.}
    \label{emb_logits}
\end{figure*}

To investigate the nodes characterized by MVRD, we select top-300 nodes most frequently involved in the queried node pairs during our experiment on Pubmed with budget $\mathcal{B}=10,000$. In Figure \ref{emb_logits} (left), we highlight these nodes in embedding spaces derived from the original node features and hidden layer of the trained GCN and LEMP4HG. Different colors denote different classes, with intensity indicating local density. We observe that both GCN and LEMP4HG form clearer class cluster than the original features. However, these nodes represented by GCN often drift into ambiguous intersection regions, while our LEMP4HG refines their representations, placing them in regions that favor correct classification. Furthermore, Figure \ref{emb_logits} (right) presents the distribution of normalized prediction logits for correct labels of these nodes with kernel density estimation, indicating that LEMP4HG notably enhances classification on these nodes.

\subsubsection{Message Synthesis}
\begin{figure*}
	\centering
	\begin{subfigure}{0.48\textwidth}
		\centering
		\includegraphics[width=\textwidth]{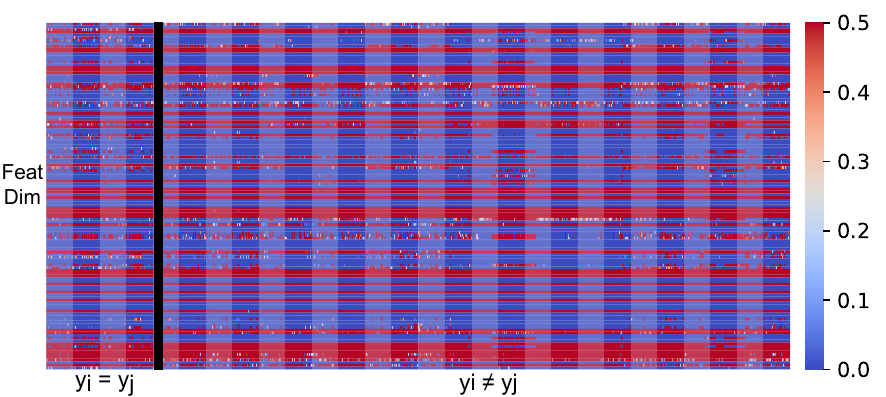}
		\caption{Gate vector that balances the contribution of source and target node embeddings.}
		\label{fig:gate}
	\end{subfigure}
	\hfill
	\begin{subfigure}{0.50\textwidth}
		\centering
		\includegraphics[width=\textwidth]{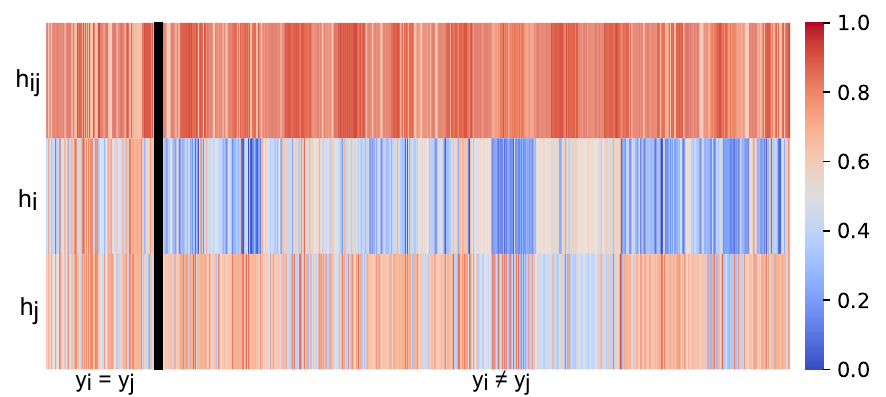}
		\caption{Similarity matrix between synthesized message $\boldsymbol{m}_{ij}$ and preliminary one $\boldsymbol{h}_{ij}$, source and target node embedding $\boldsymbol{h}_i$, $\boldsymbol{h}_j$.}
		\label{fig:sim_matrix}
	\end{subfigure}
	\caption{Gate vector analysis and similarity matrix visualization.}
	\label{gate_sim}
\end{figure*}

We analyze the message synthesis of dataset Cornell with discrimination on paired node labels ($y_i=y_j$ or $y_i\neq y_j$). In Fig. \ref{fig:gate}, we visualize the gate vector $\boldsymbol{\alpha}_{ij}^1$ in Equation \ref{gate} with $l=1$. It demonstrates that source $\boldsymbol{h}_i$ and target $\boldsymbol{h}_j$ node embedding are regularly integrated into preliminary message $\boldsymbol{h}_{ij}$ in the dimensional-level, potentially aligning with semantic structure of LM-generated connection analysis. Then, we illustrate the cosine similarity between synthesized message $\boldsymbol{m}_{ij}$ and preliminary message $\boldsymbol{h}_{ij}$, source nodes features $\boldsymbol{h}_i$ and target ones $\boldsymbol{h}_j$ in Fig. \ref{fig:sim_matrix} . We observe that preliminary message $\boldsymbol{h}_{ij}$ consistently contributes most, while the source node embedding $\boldsymbol{h}_i$ contributes more in homophilic regions ($y_i=y_j$) than heterophilic ones ($y_i\neq y_j$), conforming that message passing from source node to target one benefits from homophily.

\subsection{Interpretability}
In fact, some heterophily-specific GNNs address this issue by editing graph structures—adding homophilic edges or removing heterphilic ones. 
From our perspective, these methods represent compromises of traditional message-passing mechanisms when applied to heterophilic scenarios. 
In real-world contexts, however, every interaction has the potential to contribute valuable information for decision. The message-passing process between heterophilous node pairs is not inherently flawed;  rather, the fundamental limitation stems from traditional message-passing mechanisms' reliance on propagating a node's own features as the message. A superior message representation should emerge as the product of bidirectional information interaction and reasoning between nodes, which can be realized by language model. The integrated corpus in the reasoning process serve as augmented data that facilitate bridging the semantic gap between heterophilic nodes.

\section{Applicability and Cost Estimation}
\label{time_cost}
\subsection{Real-World Applicability}
LEMP4HG is suited for the scenarios as our budget allocation guidelines presented in Section \ref{budget}: heterophilic graphs, homophilic graphs with malignant message passing, and small to medium-sized homophilic graphs with benign message passing. For large homophilic graphs with benign message passing, some lightweight GNNs are already effective, making further enhancement unnecessary.
\paragraph{For extremely large-scale graphs} Any enhancement based on LM inference is hard to scale. Our heuristic-guided active learning strategy for selective message enhancement has reduced the cost while retaining the benefits of semantic augmentation.
\paragraph{For non-text-attributed graphs} Our LEMP4HG can work as long as informative textual representations can be constructed, as demonstrated by the tolokers dataset with numerical and categorical features converted via templates. Generalization to domains like molecular graphs remains a promising future direction.

\subsection{Efficiency Analysis}
In graph learning, edge-level complexity---which typically scales as O(E)---is often impractical for large graphs due to the sheer number of edges involved. To address this, we propose a heuristic-guided active learning strategy for edge-level message representation enhancement with LM, which selectively queries the paired nodes suffer most from message passing (i.e. most informative edges). This significantly improves feasibility and scalability of integrating LM signals into graph learning. In contrast to prior works such as LLM4HeG, which rely on LLM fine-tuning and LLM-to-SLM distillation---both of which are resource-intensive---our LEMP4HG framework adopts a lightweight, API-based querying approach. This strategy reduces both computational overhead and deployment costs, \textbf{enabling immediate integration of LM capabilities without heavy training pipelines}.

\subsection{Time Complexity Analysis}
The time complexity of LEMP4HG consists of four main parts: (1) computing the heuristic MVRD using weighted and weight-free models for edge selection; (2) querying the LM via API calls for connection analysis of the top-k node pairs with highest MVRD; (3) encoding the LM-generated analysis into textual embeddings using a fine-tuned SLM; (4) integrating these embeddings into the LM-enhanced message passing mechanism.
\subsubsection{Heuristic Calculation and Edge Selection}(1) Semi-supervised clustering: the time complexity is $O(n\cdot k\cdot d\cdot\text{iter})\approx O(n)$ as it typically the case that $n\gg d> k> iter$, where $n$ is the number of nodes, $k$ is the number of clusters, $d$ is the embedding dimension, and iter is the number of iterations. (2) Reliable difference (RD) computation: $O(m\cdot d)\approx O(m)$ for computing pairwise distances across $m$ edges. (3) Variation and modulation (VRD/MVRD) computation: $O(E)$ for simple arithmetic operations per edge; (4) Edge selection: We select top-$k$ edges with the highest MVRD scores from $m$ candidate edges approximately (the initial candidate set includes all the edges, enhanced ones are removed out every $\mathcal{I}$ epochs) by heap-based selection, resulting in a time complexity of $O(m\cdot\log k)$.

\subsubsection{Query LM for Connection Analysis} This process involves prompt construction, LM inference, response retrieval, and parsing. The overall latency is primarily influenced by the query batch size $k$ and the API rate limits under chat mode, including the maximum query rate $R_q$ (QPM, queries per minute) and maximum token rate $R_t$ (TPM, tokens per minute). For Qwen-turbo in our setup, $R_q = 60$ QPM and $R_t = 1,\!000,\!000$ TPM. To mitigate the latency, we employ asynchronous and concurrent processing strategies to improve efficiency. Alternatively, batch mode querying removes rate limits and is better suited for large-scale datasets, though it often exhibits unstable latency.

\subsubsection{Textual Encoding} The time cost of encoding the LM-generated connection analysis is primarily influenced by the encoder model size and the volume of text in the batch. In our implementation, we use a finetuned DeBERTa-base \cite{he2021deberta} with 129 million parameters, which offers a favorable trade-off between efficiency and representation capacity.

\subsubsection{LM-enhanced Message Passing}The additional computational cost compared to GCN backbone arises in the discriminative message synthesis stage. For each selected node pair, a gating function is applied over the concatenation of node textual embeddings and preliminary messages, involving a matrix-vector multiplication with complexity $O(d^2)$. This leads to a time complexity of $O(\lceil\frac{n_e}{\mathcal{I}}\rceil\cdot k\cdot d^2)$ per layer, where $n_e$ is the current training epoch, $\mathcal{I}$ is the epoch interval for querying LM, $k$ is the batch size for query, and $d$ is the hidden dimension. The message aggregation step retains the standard GCN cost of $O(m \cdot d)$, where $m$ is the number of edges.

In summary, querying LM for connection analysis is the dominant source of runtime overhead in our method. Ignoring other time-consuming components, we can estimate a lower bound on the total runtime in chat mode using the maximum query rate limit (QPM). Specifically, given $\text{budget}=\mathcal{B}$ and $\text{QPM}=R_q$, the theoretical lower bound on runtime is approximately $\frac{\mathcal{B}}{R_q}$ minutes. If batch mode is adopted, the runtime becomes highly dependent on server-side conditions and is thus difficult to estimate the runtime. Nonetheless, our empirical observations suggest that with a large query batch size (e.g. $\mathcal{B}=1000$), batch mode typically results in reduced runtime.
To further improve efficiency, several strategies can be considered: (1) deploying a lightweight LM locally; (2) using API services with more relaxed concurrency limits (e.g. Qwen-plus with $R_q=600\,\text{QPM}$); (3) or adopting API without explicit concurrency constraints (e.g. Deepseek-v3).

\subsection{Memory Analysis} 
Unlike GCN, which only maintains node-level representations with a space complexity of $O(n \cdot d)$, our method additionally stores edge-level LM-enhanced messages, incurring an extra memory cost of $O(\frac{n_e}{\mathcal{I}} \cdot k\cdot d)$ at training epoch $n_e$. While this design enables more expressive and informative message representations, it also increases overall memory footprint, particularly for dense graphs.

\subsection{Financial Cost Analysis} 
We adopt Qwen-turbo as LM to generate connection analysis between selected node pairs. According to the pricing scheme by API calls, the model incurs a cost of \$0.02 per million tokens for input (prompt) and \$0.04 per million tokens for output (completion). To estimate the financial cost, we report the average number of input and output tokens per query across 16 datasets in Table \ref{financial_cost}, using the DeBERTa-base tokenizer. Token consumption varies notably across datasets due to differences in node description length and prompt structure. For example, wikics has the highest average input length (3,195 tokens), resulting in a cost of \$0.76 per 10{,}000 queries, while lightweight datasets such as Amazon and Fitness require less than \$0.15 for the same number of queries. Despite such variations in input size, the average output length remains relatively stable (around 160--190 tokens). 

\begin{table*}[htbp]
\centering
\setlength{\tabcolsep}{3.8pt}
\caption{Statistics of token usage and associated costs. ``prompt" and ``completion" refer to the average token counts per query, and ``cost" denotes the estimated cost (USD) for 10,000 queries.}
\label{financial_cost}
\begin{tabular}{c*{16}{c}}
\toprule
 & Cornell & Texas & Wash. & Wis. & arxiv23 & Child & Amazon & Pubmed & History & Cora & citeseer & Photo & Comp. & Fitness & wikics & tolokers \\
\midrule
prompt & 1240 & 1024 & 1081 & 1439 & 700 & 672 & 185 & 905 & 667 & 507 & 560 & 712 & 412 & 211 & 3195 & 325 \\
completion & 175 & 176 & 173 & 176 & 176 & 167 & 191 & 171 & 189 & 164 & 163 & 145 & 147 & 182 & 158 & 163 \\
cost (\$) & 0.32 & 0.30 & 0.31 & 0.36 & 0.26 & 0.25 & 0.11 & 0.28 & 0.27 & 0.23 & 0.24 & 0.25 & 0.20 & 0.15 & 0.76 & 0.16 \\
\bottomrule
\end{tabular}
\end{table*}

\subsection{Case Study}
We conduct a case study to quantify the time and financial cost of LEMP4HG under different LM query budgets, and examine the corresponding accuracy trade-offs. Using the Pubmed dataset as a representative example, we report detailed cost–accuracy analyses in Table \ref{tab:cost-acc}. Specifically, to obtain edge-level connection analysis, we use the Qwen-turbo API in chat mode, with a cost of \$0.04 per million input tokens and \$0.08 per million output tokens. We simulate two realistic usage scenarios:

\paragraph{One-off scenario} Each experiment assumes no prior LM-generated connection analysis. All selected edge pairs are queried from scratch under given budget, simulating a cold-start setting.

\paragraph{Incremental scenario} Previously queried connection analyses are reused when the dataset grows or the budget increases. Only newly selected edge pairs are queried, reflecting a more practical and cost-efficient usage in dynamic or expanding graphs.

%
%

\begin{table}[t]
	\caption{Cost–accuracy trade-off of LEMP4HG on Pubmed. Costs are measured in USD and time in seconds.}
	\label{tab:cost-acc}
	\centering
	\small
	\renewcommand{\arraystretch}{1.15}
	\begin{tabular}{c|ccc|ccc}
		\toprule
		& \multicolumn{3}{c|}{One-off} & \multicolumn{3}{c}{Incremental} \\
		\cmidrule(lr){2-4} \cmidrule(lr){5-7}
		Budget 
		& Cost  & Time & Acc 
		& Cost & Time & Acc \\
		\midrule
		0     & 0    & 4.84   & 0.9346 & 0    & 4.45  & 0.9346 \\
		500   & 0.02 & 201.04 & 0.9350 & 0.01 & 151.28& 0.9350 \\
		1000  & 0.03 & 330.11 & 0.9381 & 0.01 & 107.85& 0.9381 \\
		2000  & 0.06 & 636.93 & 0.9412 & 0.01 & 156.04& 0.9412 \\
		3000  & 0.09 &1110.56 & 0.9441 & 0.01 & 155.92& 0.9441 \\
		5000  & 0.15 &1326.72 & 0.9457 & 0.02 & 311.23& 0.9457 \\
		7000  & 0.21 &1764.09 & 0.9470 & 0.02 & 308.23& 0.9470 \\
		10000 & 0.30 &2784.07 & 0.9485 & 0.04 & 482.03& 0.9485 \\
		\bottomrule
	\end{tabular}
\end{table}

\section{Prompt Design}
\label{prompt}
\subsection{Prompt for Connection Analysis}
We query the LM to analyze connections between paired nodes by providing their associated textual content, which typically describes basic entity information (e.g., paper titles and abstracts, or product descriptions and reviews). Detailed descriptions of node texts for each dataset are provided in Section~\ref{dataset_content}. The prompt templates used for different datasets are summarized in Tables~\ref{base_i} and~\ref{base_ii}. Notably, datasets within the same category share a unified prompt template, including the Acad Webpage datasets (Cornell, Texas, Washington, and Wisconsin) and the CS Citation datasets (Cora, Citeseer, and Arxiv23).

\begin{table}
	\caption{Prompt templates for querying connection analysis by the language model (Part 1).}
	\label{base_i}
	\centering
	\footnotesize
	\begin{tabularx}{\linewidth}{X}
		\toprule
		
		\textbf{Academic Webpage.} 
		Analyze the hyperlink relationship between Webpage A and Webpage B of the computer science department of a university, based on their contents provided below.
		\textcolor{gray}{\textbackslash n \textbackslash n}
		Your response should:
		\textcolor{gray}{\textbackslash n}
		1. Summarize the key content of both webpages and any notable features.
		\textcolor{gray}{\textbackslash n}
		2. Clearly explain the intellectual connection or relevance between the two webpages, highlighting how they might be related.
		\textcolor{gray}{\textbackslash n}
		3. Keep the response concise (within 200 words) and emphasize the connection between the two webpages.
		\textcolor{gray}{\textbackslash n}
		4. Use the following sentence structure: 
		``The relational implications between [Webpage A] and [Webpage B] are as below.''
		\textcolor{gray}{\textbackslash n \textbackslash n}
		Webpage A: \textcolor{blue}{$<$content A$>$}. 
		Webpage B: \textcolor{blue}{$<$content B$>$}.
		\\ \midrule
		
		\textbf{CS Citation.}
		Analyze the citation relationship between Paper A and Paper B in the field of computer science, based on their titles and abstracts provided below.
		\textcolor{gray}{\textbackslash n \textbackslash n}
		Your response should:
		\textcolor{gray}{\textbackslash n}
		1. Summarize the key content of both papers, focusing on their research questions, methods, and contributions.
		\textcolor{gray}{\textbackslash n}
		2. Clearly explain the intellectual connection or relevance between the two papers.
		\textcolor{gray}{\textbackslash n}
		3. Keep the response concise (within 200 words).
		\textcolor{gray}{\textbackslash n}
		4. Use the following sentence structure:
		``The relational implications between [Paper A] and [Paper B] are as below.''
		\textcolor{gray}{\textbackslash n \textbackslash n}
		Paper A: \textcolor{blue}{$<$content A$>$}. 
		Paper B: \textcolor{blue}{$<$content B$>$}.
		\\ \midrule
		
		\textbf{PubMed.}
		Analyze the citation relationship between Paper A and Paper B in the field of medical research on diabetes, based on their titles and abstracts provided below.
		\textcolor{gray}{\textbackslash n \textbackslash n}
		Your response should:
		\textcolor{gray}{\textbackslash n}
		1. Summarize the key content of both papers.
		\textcolor{gray}{\textbackslash n}
		2. Clearly explain the intellectual connection or relevance between the two papers.
		\textcolor{gray}{\textbackslash n}
		3. Keep the response concise (within 200 words).
		\textcolor{gray}{\textbackslash n}
		4. Use the following sentence structure:
		``The relational implications between [Paper A] and [Paper B] are as below.''
		\textcolor{gray}{\textbackslash n \textbackslash n}
		Paper A: \textcolor{blue}{$<$content A$>$}. 
		Paper B: \textcolor{blue}{$<$content B$>$}.
		\\ \midrule
		
		\textbf{History \& Children.}
		Analyze the co-purchased or co-viewed relationship between two History- or Children-related books on Amazon based on their titles and descriptions provided below.
		\textcolor{gray}{\textbackslash n \textbackslash n}
		Your response should:
		\textcolor{gray}{\textbackslash n}
		1. Summarize the main points of both items.
		\textcolor{gray}{\textbackslash n}
		2. Clearly explain the relationship between the two books.
		\textcolor{gray}{\textbackslash n}
		3. Keep the response concise (within 200 words).
		\textcolor{gray}{\textbackslash n}
		4. Use the following sentence structure:
		``The relational implications between [Book A] and [Book B] are as below.''
		\textcolor{gray}{\textbackslash n \textbackslash n}
		Book A: \textcolor{blue}{$<$content A$>$}. 
		Book B: \textcolor{blue}{$<$content B$>$}.
		\\ \midrule
		
		\textbf{Photo \& Computers.}
		Analyze the co-purchased or co-viewed relationship between two Photo- or Computers-related items on Amazon based on their user reviews provided below.
		\textcolor{gray}{\textbackslash n \textbackslash n}
		Your response should:
		\textcolor{gray}{\textbackslash n}
		1. Summarize the main points of both items' reviews.
		\textcolor{gray}{\textbackslash n}
		2. Clearly explain the relationship between the two items.
		\textcolor{gray}{\textbackslash n}
		3. Keep the response concise (within 200 words).
		\textcolor{gray}{\textbackslash n}
		4. Use the following sentence structure:
		``The relational implications between [Item A] and [Item B] are as below.''
		\textcolor{gray}{\textbackslash n \textbackslash n}
		Item A: \textcolor{blue}{$<$content A$>$}. 
		Item B: \textcolor{blue}{$<$content B$>$}.
		\\ \midrule
		
		\textbf{WikiCS.}
		Analyze the hyperlink relationship between two Wikipedia entries based on their titles and contents provided below.
		\textcolor{gray}{\textbackslash n \textbackslash n}
		Your response should:
		\textcolor{gray}{\textbackslash n}
		1. Summarize the main points of both entries.
		\textcolor{gray}{\textbackslash n}
		2. Clearly explain the relationship between the two entries.
		\textcolor{gray}{\textbackslash n}
		3. Keep the response concise (within 200 words).
		\textcolor{gray}{\textbackslash n}
		4. Use the following sentence structure:
		``The relational implications between [Entry A] and [Entry B] are as below.''
		\textcolor{gray}{\textbackslash n \textbackslash n}
		Entry A: \textcolor{blue}{$<$content A$>$}. 
		Entry B: \textcolor{blue}{$<$content B$>$}.
		\\ \midrule
		
		\textbf{Tolokers.}
		Analyze the co-work relationship between two tolokers (workers) based on their profile information and task performance statistics provided below.
		\textcolor{gray}{\textbackslash n \textbackslash n}
		Your response should:
		\textcolor{gray}{\textbackslash n}
		1. Summarize the main points of both workers' profiles and performance.
		\textcolor{gray}{\textbackslash n}
		2. Clearly explain the relationship or relevance between the two workers.
		\textcolor{gray}{\textbackslash n}
		3. Keep the response concise (within 200 words) and emphasize the relationship between the two workers.
		\textcolor{gray}{\textbackslash n}
		4. Use the following sentence structure:
		``The relational implications between [Toloker A] and [Toloker B] are as below.''
		\textcolor{gray}{\textbackslash n \textbackslash n}
		Toloker A: \textcolor{blue}{$<$content A$>$}. 
		Toloker B: \textcolor{blue}{$<$content B$>$}.
		\\ \midrule
		
		\textbf{Amazon.}
		Analyze the relationship between two items sold on Amazon based on their item names.
		The items may include products such as books, music CDs, DVDs, or VHS tapes.
		\textcolor{gray}{\textbackslash n \textbackslash n}
		Your response should:
		\textcolor{gray}{\textbackslash n}
		1. Describe and summarize the main characteristics of both items.
		\textcolor{gray}{\textbackslash n}
		2. Clearly explain the co-purchased or co-viewed relationship between the two items.
		\textcolor{gray}{\textbackslash n}
		3. Keep the response concise (within 200 words) and emphasize the relationship between the two items.
		\textcolor{gray}{\textbackslash n}
		4. Use the following sentence structure:
		``The relational implications between [Item A] and [Item B] are as below.''
		\textcolor{gray}{\textbackslash n \textbackslash n}
		Item A: \textcolor{blue}{$<$content A$>$}. 
		Item B: \textcolor{blue}{$<$content B$>$}.
		\\
		\bottomrule
	\end{tabularx}
\end{table}

\subsection{Prompt for Prediction and Explanation}
\label{PE}
Following TAPE~\cite{he2023harnessing}, we query the LM for category prediction and explanation to enhance node representations. Unlike TAPE, which focuses on titles and abstracts in citation networks, we adopt a more general formulation by defining the LM input as the basic entity information specific to each dataset, such as titles and abstracts for citation graphs and product descriptions for e-commerce graphs. The corresponding prompt templates are provided in Tables~\ref{BIPE_i} and~\ref{BIPE_ii}.
\subsection{Textual Generation}
We present an illustrative example of the connection analysis generated by the language model for a node pair from the PubMed dataset.

\textit{“The relational implications between Paper A and Paper B are as below. Paper A investigates the association between type 2 diabetes and osteopenia, finding variable BMD outcomes depending on age and gender. It emphasizes the need for individualized assessment. Paper B reviews the role of insulin as an anabolic agent in bone, suggesting that insulin may influence bone density and strength in diabetic patients. The intellectual connection lies in their shared focus on diabetes and bone health, with Paper A providing empirical data and Paper B offering a mechanistic explanation involving insulin's role. Together, they highlight the complex relationship between diabetes and bone metabolism, with Paper B potentially explaining the findings observed in Paper A."}

\begin{table}
    \caption{Prompt templates for querying connection analysis by the language model (Part 2).}
    \label{base_ii}
    \centering
    \footnotesize
    \begin{tabularx}{\linewidth}{X}
        \toprule

        \textbf{Fitness.}
        Analyze the co-purchased or co-viewed relationship between two fitness-related items sold on Amazon based on their item titles provided below.
        \textcolor{gray}{\textbackslash n \textbackslash n}
        Your response should:
        \textcolor{gray}{\textbackslash n}
        1. Describe and summarize the main points of both items.
        \textcolor{gray}{\textbackslash n}
        2. Clearly explain the relationship or relevance between the two items.
        \textcolor{gray}{\textbackslash n}
        3. Keep the response concise (within 200 words) and emphasize the relationship between the two items.
        \textcolor{gray}{\textbackslash n}
        4. Use the following sentence structure:
        ``The relational implications between [Item A] and [Item B] are as below.''
        \textcolor{gray}{\textbackslash n \textbackslash n}
        Item A: \textcolor{blue}{$<$content A$>$}. 
        Item B: \textcolor{blue}{$<$content B$>$}.
        \\ \midrule

        \textbf{Products.}
        Analyze the co-purchased relationship between two items sold on Amazon based on their product descriptions provided below.
        \textcolor{gray}{\textbackslash n \textbackslash n}
        Your response should:
        \textcolor{gray}{\textbackslash n}
        1. Summarize the main points of both products' descriptions.
        \textcolor{gray}{\textbackslash n}
        2. Clearly explain the relationship or relevance between the two items.
        \textcolor{gray}{\textbackslash n}
        3. Keep the response concise (within 200 words) and emphasize the relationship between the two items.
        \textcolor{gray}{\textbackslash n}
        4. Use the following sentence structure:
        ``The relational implications between [Item A] and [Item B] are as below.''
        \textcolor{gray}{\textbackslash n \textbackslash n}
        Item A: \textcolor{blue}{$<$content A$>$}. 
        Item B: \textcolor{blue}{$<$content B$>$}.
        \\

        \bottomrule
    \end{tabularx}
\end{table}

\begin{table}
    \caption{Prompt templates for querying category prediction and explanation by the language model (Part 1).}
    \label{BIPE_i}
    \centering
    \footnotesize
    \begin{tabularx}{\linewidth}{X}
        \toprule

        \textbf{Academic Webpage.}
        [Webpage Content]: \textcolor{blue}{$<$content$>$}
        \textcolor{gray}{\textbackslash n \textbackslash n}
        [Question]: Which of the following categories does this webpage belong to: Student, Project, Course, Staff, Faculty?
        If multiple options apply, provide a comma-separated list ordered from most to least related, and for each selected category, explain how it is reflected in the text.
        \textcolor{gray}{\textbackslash n \textbackslash n}
        [Answer]:
        \\ \midrule

        \textbf{PubMed.}
        [Paper Info]: \textcolor{blue}{$<$content$>$}
        \textcolor{gray}{\textbackslash n \textbackslash n}
        [Question]: Does the paper involve any cases of Type~1 diabetes, Type~2 diabetes, or experimentally induced diabetes?
        Please provide one or more answers from the following options: Type~1 diabetes, Type~2 diabetes, Experimentally induced diabetes.
        If multiple options apply, give a comma-separated list ordered from most to least related, and for each selected option, provide a detailed explanation with supporting quotes from the text.
        \textcolor{gray}{\textbackslash n \textbackslash n}
        [Answer]:
        \\ \midrule

        \textbf{arxiv23.}
        [Paper Info]: \textcolor{blue}{$<$content$>$}
        \textcolor{gray}{\textbackslash n \textbackslash n}
        [Question]: Which arXiv CS subcategory does this paper belong to?
        Provide five likely arXiv CS subcategories as a comma-separated list ordered from most to least likely, using the format ``cs.XX'', and briefly justify your choices.
        \textcolor{gray}{\textbackslash n \textbackslash n}
        [Answer]:
        \\ \midrule

        \textbf{History.}
        [Book Info]: \textcolor{blue}{$<$content$>$}
        \textcolor{gray}{\textbackslash n \textbackslash n}
        [Question]: Which of the following History subcategories does this book belong to:
        World, Americas, Asia, Military, Europe, Russia, Africa, Ancient Civilizations, Middle East, Historical Study \& Educational Resources, Australia \& Oceania, Arctic \& Antarctica?
        If multiple options apply, provide a comma-separated list ordered from most to least related, and explain how each category is reflected in the text.
        \textcolor{gray}{\textbackslash n \textbackslash n}
        [Answer]:
        \\ \midrule
        
        \textbf{Children.}
        [Book Info]: \textcolor{blue}{$<$content$>$}
        \textcolor{gray}{\textbackslash n \textbackslash n}
        [Question]: Which of the following sub-categories of Children does this book belong to:
        Literature \& Fiction, Animals, Growing Up \& Facts of Life, Humor, Cars Trains \& Things That Go, Fairy Tales Folk Tales \& Myths, Activities Crafts \& Games, Science Fiction \& Fantasy, Classics, Mysteries \& Detectives, Action \& Adventure, Geography \& Cultures, Education \& Reference, Arts Music \& Photography, Holidays \& Celebrations, Science Nature \& How It Works, Early Learning, Biographies, History, Children's Cookbooks, Religions, Sports \& Outdoors, Comics \& Graphic Novels, Computers \& Technology?
        If multiple options apply, provide a comma-separated list ordered from most to least related, and explain how each choice is reflected in the text.
        \textcolor{gray}{\textbackslash n \textbackslash n}
        [Answer]:
        \\ \midrule

        \textbf{Cora.}
        [Paper Info]: \textcolor{blue}{$<$content$>$}
        \textcolor{gray}{\textbackslash n \textbackslash n}
        [Question]: Which of the following AI sub-categories does this paper belong to:
        Case Based, Genetic Algorithms, Neural Networks, Probabilistic Methods, Reinforcement Learning, Rule Learning, Theory?
        If multiple options apply, provide a comma-separated list ordered from most to least related, and explain how each choice is reflected in the text.
        \textcolor{gray}{\textbackslash n \textbackslash n}
        [Answer]:
        \\
        \bottomrule
    \end{tabularx}
\end{table}

\begin{table}
    \caption{Prompt templates for querying category prediction and explanation by the language model (Part 2).}
    \label{BIPE_ii}
    \centering
    \footnotesize
    \begin{tabularx}{\linewidth}{X}
        \toprule
        \textbf{Citeseer.}
        [Paper Info]: \textcolor{blue}{$<$content$>$}
        \textcolor{gray}{\textbackslash n \textbackslash n}
        [Question]: Which of the following computer science sub-categories does this paper belong to:
        Agents, Machine Learning, Information Retrieval, Database, Human-Computer Interaction, Artificial Intelligence?
        If multiple options apply, provide a comma-separated list ordered from most to least related, and explain how each choice is reflected in the text.
        \textcolor{gray}{\textbackslash n \textbackslash n}
        [Answer]:
        \\ \midrule
        
        \textbf{WikiCS.}
        [Entry Info]: \textcolor{blue}{$<$content$>$}
        \textcolor{gray}{\textbackslash n \textbackslash n}
        [Question]: Which of the following computer science sub-categories does this Wikipedia entry belong to:
        Computational Linguistics, Databases, Operating Systems, Computer Architecture, Computer Security, Internet Protocols, Computer File Systems, Distributed Computing Architecture, Web Technology, Programming Language Topics?
        If multiple options apply, provide a comma-separated list ordered from most to least related, and explain how each choice is reflected in the text.
        \textcolor{gray}{\textbackslash n \textbackslash n}
        [Answer]:
        \\ \midrule

        \textbf{Tolokers.}
        [Worker Info]: \textcolor{blue}{$<$content$>$}
        \textcolor{gray}{\textbackslash n \textbackslash n}
        [Question]: What is the probability that the worker will be banned from a specific project?
        Choose one of the following options:
        Very Low ($<10\%$), Low (10--30\%), Moderate (30--50\%), High (50--70\%), Very High ($>70\%$).
        Then explain how the selected option is supported by the text.
        \textcolor{gray}{\textbackslash n \textbackslash n}
        [Answer]:
        \\ \midrule

        \textbf{Amazon.}
        [Item Name]: \textcolor{blue}{$<$content$>$}
        \textcolor{gray}{\textbackslash n \textbackslash n}
        [Question]: The item is a product such as a book, music CD, DVD, or VHS tape.
        What rating grade is the item likely to receive?
        Choose one of the following options:
        Good (score 5--3.5), Average (score 3.5--2.5), Bad (score 2.5--1).
        Then explain how the selected option is supported by the text.
        \textcolor{gray}{\textbackslash n \textbackslash n}
        [Answer]:
        \\ \midrule

        \textbf{Photo.}
        [Item Review]: \textcolor{blue}{$<$content$>$}
        \textcolor{gray}{\textbackslash n \textbackslash n}
        [Question]: Which of the following photo sub-categories does this item belong to:
        Film Photography, Video, Digital Cameras, Accessories, Binoculars \& Scopes, Lenses, Bags \& Cases, Lighting \& Studio, Flashes, Tripods \& Monopods, Underwater Photography, Video Surveillance?
        If multiple options apply, provide a comma-separated list ordered from most to least related, and explain how each choice is reflected in the text.
        \textcolor{gray}{\textbackslash n \textbackslash n}
        [Answer]:
        \\ \midrule

        \textbf{Computers.}
        [Item Review]: \textcolor{blue}{$<$content$>$}
        \textcolor{gray}{\textbackslash n \textbackslash n}
        [Question]: Which of the following computer sub-categories does this item belong to:
        Laptop Accessories, Computer Accessories \& Peripherals, Computer Components, Data Storage, Networking Products, Monitors, Computers \& Tablets, Tablet Accessories, Servers, Tablet Replacement Parts?
        If multiple options apply, provide a comma-separated list ordered from most to least related, and explain how each choice is reflected in the text.
        \textcolor{gray}{\textbackslash n \textbackslash n}
        [Answer]:
        \\ \midrule

        \textbf{Fitness.}
        [Item Title]: \textcolor{blue}{$<$content$>$}
        \textcolor{gray}{\textbackslash n \textbackslash n}
        [Question]: Which of the following fitness sub-categories does this item belong to:
        Other Sports, Exercise \& Fitness, Hunting \& Fishing, Accessories, Leisure Sports \& Game Room, Team Sports, Boating \& Sailing, Swimming, Tennis \& Racquet Sports, Golf, Airsoft \& Paintball, Clothing, Sports Medicine?
        If multiple options apply, provide a comma-separated list ordered from most to least related, and explain how each choice is reflected in the text.
        \textcolor{gray}{\textbackslash n \textbackslash n}
        [Answer]:
        \\ \midrule

        \textbf{Products.}
        [Product Description]: \textcolor{blue}{$<$content$>$}
        \textcolor{gray}{\textbackslash n \textbackslash n}
        [Question]: Which of the following categories does this product belong to?
        Provide five likely categories as a comma-separated list ordered from most to least likely, and briefly justify your choices.
        \textcolor{gray}{\textbackslash n \textbackslash n}
        [Answer]:
        \\

        \bottomrule
    \end{tabularx}
\end{table}

\section{Related Works}

\subsection{Graph Neural Networks for Heterophily}
Existing GNNs for heterophilic graphs mainly adopt two strategies: graph structure refinement and GNN architecture refinement. The former optimizes node receptive fields by selectively expanding neighborhood and pruning unbefitting connections. For example, Geom-GCN \cite{pei2020geom} extends neighborhood by embedding-based proximity, while U-GCN \cite{jin2021universal} extract information from 1-hop, 2-hop and k-nearest neighbors simultaneously. SEGSL \cite{zou2023se} refines graph topology using structural entropy and encoding trees, while DHGR \cite{bi2024make} based on label or feature distribution similarity. GNN-SATA \cite{yang2024graph} dynamically adds or removes edges by associating topology with attributes. The latter optimizes the neighboring aggregation and representation updating functions. For example, FAGCN \cite{bo2021beyond} discriminatively aggregate low-frequency and high-frequency signals, while OGNN \cite{song2023ordered} updates node representation with multi-hop neighbors by orders. EGLD \cite{zhang2024unleashing} utilize dimension masking to balance low and high-pass filtered features. 
However, these methods typically treat text as independent node-level features, without explicitly modeling inter-node relational semantics, leaving the message representation fundamentally unchanged. In contrast, we leverage LMs to generate and encode pairwise semantic analyses, redefining the message itself rather than merely enhancing node features.

\subsection{LLMs for TAG Graph Learning}
Existing works integrating LM into graph tasks achieve great success \cite{fan2024graph, ren2024survey}, with three main methods. \cite{li2023survey} (1) LM as enhancer \cite{tan2023walklm, liu2023one}, where LM generate text and embeddings to enhance GNNs classifier. For example, TAPE \cite{he2023harnessing} improves node representations with SLM-encoded text embeddings and LM-generated explanation for classification and pseudo labels. (2) LM as predictor, where graph structures are transformed into textual descriptions \cite{zhao2023graphtext, ye2023language}, or textual features are combined with GNN-encoded structural information \cite{chai2023graphllm, tang2024graphgpt} for LM inference. (3) GNN-LM alignment, which aligns GNN and LM embeddings through contrastive learning \cite{radford2021learning}, interactive supervision \cite{zhao2022learning}, or GNN-guided LM training \cite{mavromatis2023train}. However, these methods are not tailored for heterophily. 
Although LLM4HeG \cite{wu-etal-2025-exploring} leverages LMs for edge discrimination and reweighting, it does not alter the conventional message formulation, where node embeddings are directly propagated. In contrast, we employ LM-generated relational texts to construct enhanced messages, fundamentally reshaping information transmission in heterophilic settings.

\subsection{Graph Active Learning}
Traditional graph active learning \cite{cai2017active} selects nodes and query labels to improve test performance within a limited budget. Existing researches mainly optimize selection strategy from multiple perspectives, such as the diversity and representativeness of the selected nodes \cite{zhang2021rim}. For example, Ma et al.\cite{ma2022partition} select nodes from distinct communities for broad coverage, while Zhang et al.\cite{zhang2021alg} prioritize the nodes with higher influence scores. Some approaches employ reinforcement learning to optimize model accuracy \cite{hu2020graph, zhang2022batch}. With the prevalence of LM, LLM-GNN \cite{chen2023label} enhances node selection by using LM as annotators, addressing limited ground truth and noise in annotations. 
However, these works focus on node label annotation, overlooking that unreliable interactions between node pairs can distort representation learning. In contrast, we selectively enhance edge-level semantic information, directly targeting distortion in message passing.

\section{Conclusion}

In this paper, we analyze and theoretically prove why vanilla GNNs fail to achieve good classification performance in scenarios with significant differences in class homophily, especially in graph anomaly detection. We also propose a new metric, class homophily variance, to quantitatively describe the severity of this phenomenon.
To address the issue of high class homophily variance, we introduce \texttt{HEAug}, which alleviates this problem by generating homophilic edges to augment the graph structure rather than modifying original relationships.
Experiments have demonstrated the effectiveness of our method in various scenarios including graph anomaly detection datasets, simulation and edgeless node classification, and have proven that the edges generated by the model have low class homophily variance.

\normalem
\bibliographystyle{IEEEtran}
\bibliography{IEEEabrv,tapmi}

\end{document}